\documentclass[preprint,12pt,3p,authoryear]{elsarticle}

\usepackage{graphbox,graphicx}
\usepackage{amssymb}

\usepackage{gensymb}
\usepackage[binary-units,separate-uncertainty=true,multi-part-units=single]{siunitx}
\usepackage{amsmath}
\usepackage{todonotes}
\usepackage{blkarray}
\usepackage{rotating}
\usepackage{cprotect}
\usepackage{hyperref}
\usepackage{array}
\usepackage{listings}

\journal{Remote Sensing of Environment}

\biboptions{authoryear}

\begin{document}

\begin{frontmatter}

\title{Virtual laser scanning with HELIOS++: A novel take on ray tracing-based simulation of topographic 3D laser scanning}

%%include affiliations in footnotes:
\author[1]{Lukas Winiwarter\corref{mycorrespondingauthor}}
\cortext[mycorrespondingauthor]{Corresponding authors}
\ead{lukas.winiwarter@uni-heidelberg.de}

\author[3]{Alberto Manuel Esmorís Pena}

\author[1]{Hannah Weiser}
\author[1,2]{Katharina Anders}
\author[3]{Jorge Martínez Sanchez}
\author[1]{Mark Searle}
\author[1,2]{Bernhard Höfle\corref{mycorrespondingauthor}}
\ead{hoefle@uni-heidelberg.de}

\address[1]{3DGeo Research Group, Institute of Geography, Heidelberg University, Germany}
\address[3]{Centro Singular de Investigación en Tecnoloxías Intelixentes, CiTIUS, USC, Spain}
\address[2]{Interdisciplinary Center for Scientific Computing (IWR), Heidelberg University, Germany}

\begin{abstract}
Topographic laser scanning is a remote sensing method to create detailed 3D point cloud representations of the Earth's surface. Since data acquisition is expensive, simulations can complement real data given certain premises are available: i) a model of 3D scene and scanner, ii) a model of the beam-scene interaction, simplified to a computationally feasible while physically realistic level, and iii) an application for which simulated data is fit for use. A number of laser scanning simulators for different purposes exist, which we enrich by presenting HELIOS++. HELIOS++ is an open-source simulation framework for terrestrial static, mobile, UAV-based and airborne laser scanning implemented in C++. The HELIOS++ concept provides a flexible solution for the trade-off between physical accuracy (realism) and computational complexity (runtime, memory footprint), as well as ease of use and of configuration. Unique features of HELIOS++ include the availability of Python bindings (\verb|pyhelios|) for controlling simulations, and a range of model types for 3D scene representation. Such model types include meshes, digital terrain models, point clouds and partially transmissive voxels, which are especially useful in laser scanning simulations of vegetation. In a scene, object models of different types can be combined, so that representations spanning multiple spatial scales in different resolutions and levels of detail are possible. We aim for a modular design, where the core components of platform, scene, and scanner can be individually interchanged, and easily configured by working with XML files and Python bindings.
Virtually scanned point clouds may be used for a broad range of applications. Our literature review of publications employing virtual laser scanning revealed the four categories of use cases prevailing at present: data acquisition planning, method evaluation, method training and sensing experimentation.
To enable direct interaction with 3D point cloud processing and GIS software, we support standard data formats for input models (Wavefront Objects, GeoTIFFs, ASCII xyz point clouds) and output point clouds (LAS/LAZ and ASCII).
HELIOS++ further allows the simulation of beam divergence using a subsampling strategy, and is able to create full-waveform outputs as a basis for detailed analysis. As generation and analysis of waveforms can strongly impact runtimes, the user may set the level of detail for the subsampling, or optionally disable full-waveform output altogether.
 A detailed assessment of computational considerations and a comparison of HELIOS++ to its predecessor, HELIOS, reveal reduced runtimes by up to 83~\%. At the same time, memory requirements are reduced by up to 94~\%, allowing for much larger (i.e. more complex) 3D scenes to be loaded into memory and hence to be virtually acquired by laser scanning simulation.
\end{abstract}


\end{frontmatter}

\section{Introduction}

\begin{figure} [h!]
    \centering
    \includegraphics[width=.9\linewidth]{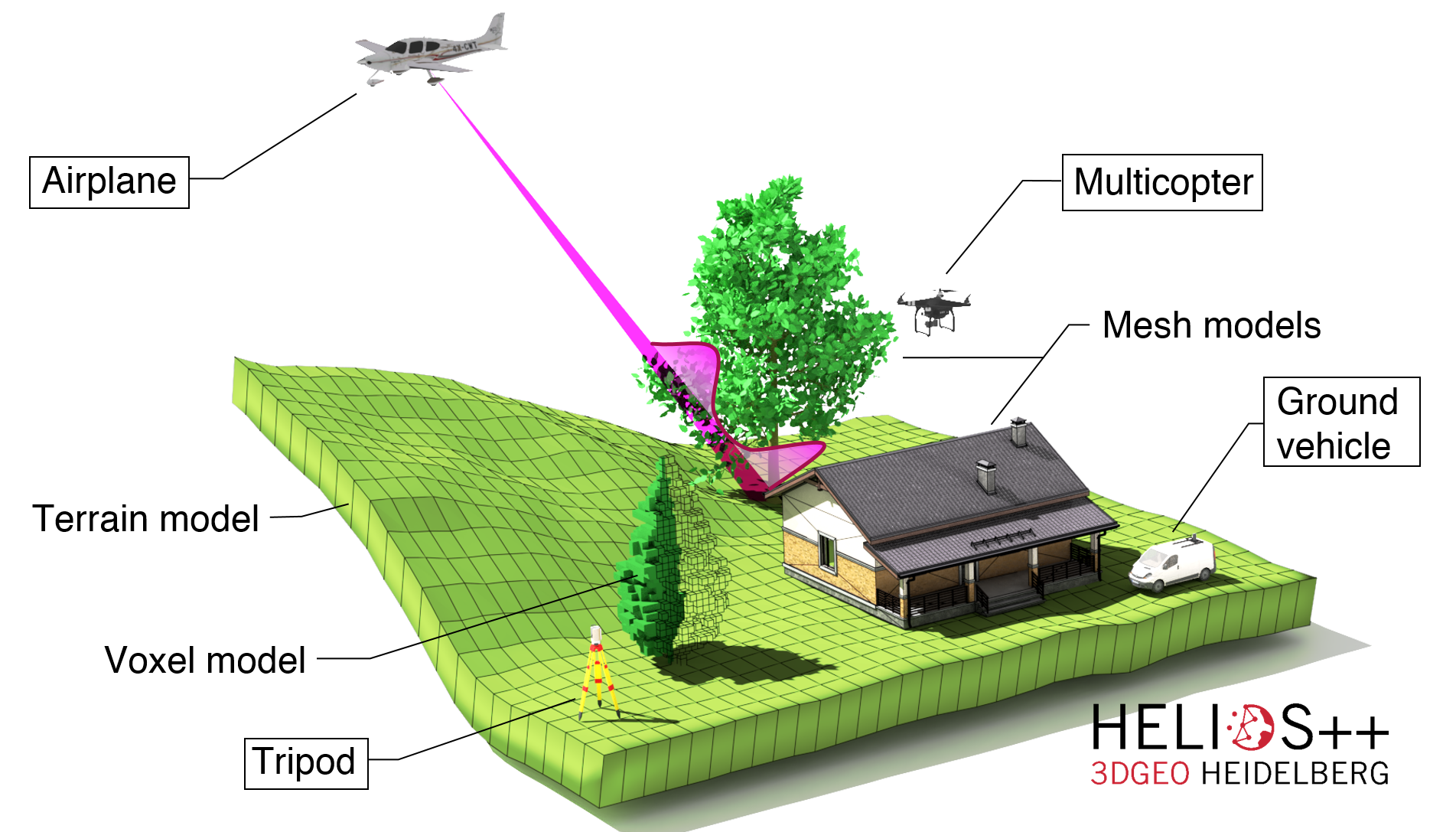}
    \caption{Schematic concept of HELIOS++, showcasing platforms (boxed labels) and object models composing a scene (non-boxed labels). A variety of model types to represent 3D scenes are supported: terrain models, voxel models (custom .vox format or XYZ point clouds) and mesh models. For platforms, four options are currently supported: airplane, multicopter, ground vehicle and static tripod. A schematic diverging laser beam and its corresponding waveform (magenta) is shown being emitted from the airplane and interacting with a mesh model tree and the rasterised ground surface.}
    \label{fig:geile_figure}
\end{figure}

Simulation of physical processes is often carried out when experiments are not feasible or simply impossible, or to find parameters that produce a certain outcome if inversion is non-trivial. In virtual laser scanning (VLS), simulations of LiDAR (Light Detection and Ranging) create 3D point clouds from models of scenes, platforms, and scanners (Figure~\ref{fig:geile_figure}), that aim to recreate real-world scenarios of laser scanning acquisitions. Such simulated point clouds may, for certain use cases, replace real data, and may even allow for analyses where real data capture is not feasible, e.g. due to technical, economical or logistic constraints, or when simulating hardware which is not yet existing. However, there are use cases where VLS is not appropriate, for example, when analysing effects only partially modelled in the simulation such as penetration of the laser into opaque objects. In a similar argument, (passive) photogrammetry is inadequate for the reconstruction of a non-textured flat area, but still a useful method for many other tasks. Therefore, VLS can be seen as a tool to acquire 3D geospatial data under certain \emph{premises}. These include:
\begin{enumerate}
    \item an adequate model of the 3D scene and the scanner, as well as the platform behaviour,
    \item a simplification of the real-world beam-scene interactions to a computationally feasible and physically realistic level, and finally,
    \item an application for which VLS data is fit for use.
\end{enumerate}

VLS can easily and cheaply produce large amounts of data with very well defined properties and known ground truth. Parameters (e.g. tree attributes such as crown base height) can be extracted from the scene model (e.g. a mesh object that is being scanned) automatically and without errors, and these parameters can in turn be used for training or validation of algorithms that attempt to extract them from point cloud data. Due to the low cost compared to real data acquisitions, VLS can be combined with Monte-Carlo simulations to solve non-continuous optimisation problems on scan settings and acquisition strategies. In a research workflow, VLS experiments may be employed to identify promising candidate settings before carrying out a selected number of real experiments that are used to answer the respective research questions.

In this paper, we present a novel take on ray tracing-based VLS covering the given premises (1-3). This is implemented as the open source software package HELIOS++ (Heidelberg LiDAR Operations Simulator ++)\footnote{HELIOS++ is available on GitHub (https://github.com/3dgeo-heidelberg/helios) and is licensed under both GNU GPL and GNU LGPL. HELIOS++ is also indexed with Zenodo \citep{lukas_winiwarter_2021_4452871}.}. HELIOS++ is the successor of HELIOS \citep{Bechtold_2016}, with a completely new code base, implemented in C++ (whereas the former version was implemented in Java). HELIOS++ improves over HELIOS in terms of memory footprint and runtime, and also in functionality, correctness of implemented algorithms, and usability, making it versatile and highly performant to end users.

We first motivate the need for a novel VLS framework by a survey of previous methodologies to laser scanning simulation and their applications, and point out the unique features of HELIOS++ (Section~\ref{sec:sota}). 
In Section~\ref{sec:implementaton}, we present the architecture and design considerations of HELIOS++. We then show different types of applications and conduct a systematic literature survey of uses of HELIOS in Section~\ref{sec:applications}. Technical considerations concerning the handling of big 3D scenes and ray tracing are dealt with in Section~\ref{sec:computational} and conclusions are drawn in Section~\ref{sec:conclusions}.

\section{Existing implementations and state of the art virtual laser scanning}
\label{sec:sota}

Simulating a process within a system always involves a simplified substitute of reality. The complexity of this substitute depends on the process understanding, on computational considerations and on the specific problem that is to be solved. Approaches with different levels of simulation complexity exist regarding i) the type of input scene model (e.g. 2.5D digital elevation models (DEMs) vs. 3D meshes) and ii) how the interaction of beam and object is modeled (e.g. single ray/echo vs. full-waveform). An overview of publications of these approaches is listed in Table~\ref{tab:vls_overview}.

\begin{table}[]
\centering
\begin{tabular}{>{\raggedright}p{0.25\columnwidth}|lllll}
\textbf{Publication}                                                                                     & \textbf{Platforms}                                                  & \textbf{\begin{tabular}[c]{@{}l@{}}Beam\\ div.\end{tabular}} & \textbf{FWF} & \textbf{Scene} & \textbf{Comments}                                                                   \\ \hline  \hline
\begin{tabular}[c]{@{}l@{}}\citet{North.1996},\\ \citet{North.2010}\end{tabular}                         & satellite                                                     & \checkmark                                                                & \checkmark             & 3D                  & FLIGHT model                                                                        \\ \hline

\citet{Lewis.1999}                                                                                       & ALS                                                                 & \checkmark                                                                & \checkmark             & 3D                  & \begin{tabular}[c]{@{}l@{}}used by\\ \citet{Calders.2013} \& \\ \citet{Disney.2010}\end{tabular} \\ \hline
\citet{Tulldahl.1999}                                                                                    & ALS                                                                 & \checkmark                                                             & \checkmark             & 3D                  & for bathymetry                                                                       \\ \hline
\citet{Ranson.2000}                                                                                      & \begin{tabular}[c]{@{}l@{}}ALS\\ (nadir)\end{tabular}               & \checkmark                                                             & \checkmark             & 3D                  &                                                                                     \\ \hline
\citet{Holmgren.2003}                                                                                    & ALS                                                                 &                                                                           &                        & 3D                  &                                                                                     \\ \hline
\citet{Goodwin.2007}                                                                                     & ALS                                                                 &                                                                           &                        & 3D                  & LITE model     
\\ \hline
\citet{Lohani.2007}                                                                                      & ALS                                                                 &                                                                           &                        & 2.5D                &                                                                                                                                                         \\ \hline
\citet{Morsdorf.2007}                                                                                    & ALS                                                                 & \checkmark                                                                & \checkmark             & 3D                  & using POVray                                                                        \\ \hline
\citet{Kim.2009}                                                                                         & ALS                                                                 &                                                                           &                        & 3D                  &                                                                                     \\ \hline
\citet{Kukko.2009}                                                                                       & ALS, MLS                                                            & \checkmark                                                                & \checkmark             & 2.5D                &                                                                                     \\ \hline
\citet{Hodge.2010}                                                                                       & TLS                                                                 & \checkmark                                                                & \checkmark             & 2.5D                &                                                                                     \\ \hline
\citet{Kim.2012}                                                                                         & ALS                                                                 & \checkmark                                                                & \checkmark             & 3D               &                                                                                     \\ \hline
\citet{Wang.July2013}                                                                                    & TLS                                                                 &                                                                           &                        & 3D                  &                                                                                     \\ \hline
\citet{GastelluEtchegorry.2015} & \begin{tabular}[c]{@{}l@{}}satellite,\\ ALS, TLS\end{tabular} & \checkmark                                                                & \checkmark             & 3D                  & DART model      \\ \hline
\citet{Bechtold_2016}                                                                                         &\begin{tabular}[c]{@{}l@{}}ALS, TLS,\\ MLS, ULS\end{tabular}                                                                & \checkmark                                                                & \checkmark             & 3D               &      HELIOS                                                                               \\ 
\end{tabular}
\caption{Overview of virtual laser scanning simulators and associated publications. For each simulator, a check mark (\checkmark) is added if they support simulation of finite (non-zero) beam divergence ("Beam div.") and full waveforms ("FWF"). Scene representation may be in full 3D or 2.5D, i.e. raster-based.}
\label{tab:vls_overview}
\end{table}

A simple airborne laser scanning (ALS) simulator is presented by \citet{Lohani.2007} for use in research and education. Their tool comes with a user-friendly graphical user interface (GUI) and allows selecting different scanner and trajectory configurations. This simulator models the laser ray as an infinitesimal beam with zero divergence to simplify the ray tracing procedure. The scene is represented by a 2.5D elevation raster, which allows only a simplified representation of the Earth's surface.

In a different application, the interactions between laser beams and forest canopies are investigated by \citet{Goodwin.2007}, \citet{Holmgren.2003} and \citet{Lovell.2005}. They present approaches of combining 3D forest modelling and ALS simulation using ray tracing. As in \citet{Lohani.2007}, their simulators all model the laser beam as an infinite straight line, which intersects the scene in one distinct point. 

Full-waveform laser scanners can record the full waveform of the backscattered signal, providing information about the objects in a scene that are illuminated by the conic beam. This waveform from a finite footprint was specifically simulated by \citet{Ranson.2000} in the forestry context and by \citet{Tulldahl.1999} for airborne LiDAR bathymetry. \citet{Morsdorf.2007} use the ray tracing software \textit{POV-Ray}\footnote{http://www.povray.org/} to model the waveform of laser scans of a 3D tree model from a combination of intensity and depth images.

\citet{Kukko.2009} aim for a more complete and universal LiDAR simulator that considers platform and beam orientation, pulse transmission depending on power distribution and laser beam divergence, beam interaction with the scene, and full-waveform recording. Their approach models the physics involved in LiDAR measurements in high detail, and is demonstrated for a use case in forestry. Similar to the work by \citet{Lohani.2007}, they use 2.5D elevation maps to represent the scene. This makes the simulator useful for airborne simulations over terrain or building models, where the scene is scanned from above and scene elements are assumed to be solid and opaque. However, it is less suited for penetrable 3D objects such as vegetation or overhanging geometries, especially for the simulation of ground-based acquisitions which are less in a bird's-eye view perspective. \citet{Kim.2012} present a similarly detailed simulator, which includes radiometric simulation and recording of the waveform and includes explicit 3D object representations. It is unclear if it also supports static platforms such as TLS.

TLS simulations are the sole focus of some studies for specific applications, such as leaf area index inversion \citep{Wang.July2013} or TLS measurement error quantification \citep{Hodge.2010}. While \citet{Wang.July2013} use a more simple model assuming no beam divergence, the simulation described by \citet{Hodge.2010} includes both the modelling of beam divergence and recording of the waveform. Their simulation again uses 2.5D elevation models to represent the scene, which is appropriate for their particular objective of error quantification in high-resolution, short-range TLS of natural surfaces, specifically fluvial sediment deposits, of small scenes (area of \SI{1x1}{\metre}).

Established Monte-Carlo ray tracing simulator are used and being extended for airborne and satellite laser scanning simulations. The \textit{librat} model \citep{Calders.2013, Disney.2009, Disney.2010}, a modular development of ARARAT \citep{Lewis.1993} is such a Monte-Carlo simulator. Similarly, \citet{North.2010} extend the 3D radiative transfer model FLIGHT \citep{North.1996} to model satellite LiDAR waveforms. Monte-Carlo methods represent a simple, robust and versatile set of techniques to solve multi-dimensional problems by repeatedly sampling from a probability density function describing the system that is investigated \citep{Disney.2000}. These stochastic methods are useful for simulating multi-scattering processes, e.g. for modeling canopy reflectance. The main drawback of Monte-Carlo ray tracing methods are high computation times to simulate sufficient photons for the scattering model to converge to an accurate solution \citep{Disney.2000, GastelluEtchegorry.2016}. The LiDAR extension of the Discrete Anisotropic Radiative Transfer (DART) model attempts to alleviate these restrictions by quickly selecting scattering directions of simulated photons using the so-called Box method and modeling their propagation and interaction using a Ray Carlo method, which combines classical Monte-Carlo and ray tracing methods \citep{GastelluEtchegorry.2016}. 

A comprehensive review of simulators for the generation of point cloud data, including LiDAR simulators, is presented in \citet{Schlager_2020} with focus on their applicability to generate data in the context of driver assistance systems and autonomous driving vehicles. In this context, they analyse algorithms with respect to their fidelity, operating principles, considered effects and possible improvements.

In contrast to most of the previously mentioned approaches, HELIOS++ provides a framework for full 3D laser scanning simulation with multiple platforms (terrestrial (TLS), mobile (MLS), UAV-borne (ULS) and airborne (ALS)), and a flexible system to represent scenes, which allows combination of input data from multiple sources and data formats (Figure~\ref{fig:geile_figure}). The simulation of beam divergence and full waveform recording are supported. While HELIOS++ may not be as realistic in terms of physical accuracy regarding the energy budget of a single laser shot as, e.g., DART, it provides a sensible trade-off between computational efforts and resulting point cloud quality. Users can simulate VLS over a large range of scales, and even combine different scales in one scene. For example, a highly detailed tree model with individual leaves might be placed in a forest scene represented by (transmissive) voxels \citep{Weiser2020}, while using a rasterised digital terrain model as ground surface. This allows to model the influence of the surrounding of a particular object of interest on the derived VLS point clouds. Furthermore, HELIOS++ aims for high usability by providing a comprehensible set of parameters, while not overwhelming the users with options. These parameters represent the state of the art and are supported by peer-reviewed literature. Since HELIOS++ can be used from the command line and from within Python, workflows integrating HELIOS++ can be easily scripted and automated, as well as linked to external software (e.g. GIS, 3D point cloud processing software, Jupyter Notebooks, and others).

HELIOS++ comes with an extensive documentation of all algorithms that are used in the simulations, including examples and references to the relevant literature describing the implemented methods. Furthermore, the open source implementation allows any user to i) inspect and ii) alter/adapt the source code of the program. For ease of use, we provide pre-compiled versions that are ready to use for major operating systems (Microsoft Windows 10 and Debian Linux 10.7 Buster), and the option to use Python as a scripting language to create, manipulate and simulate VLS data acquisition with HELIOS++.

\section{Implementation of HELIOS++}
\label{sec:implementaton}
This section introduces the concepts and interfaces of HELIOS++, for which the important components are the overall architecture and modules (Section~\ref{sec:architecture}), platforms (Section~\ref{sec:platforms}), scanners and laser beam deflectors (Section~\ref{sec:scanners}), the waveform simulation (Section~\ref{sec:fullwave}), input formats (Section~\ref{sec:object_models}) and output formats (Section~\ref{sec:output}), and the aspect of randomness and repeatability of results (Section~\ref{sec:randomness}).

\subsection{Architecture and modules}
\label{sec:architecture}
\begin{figure}
    \centering
    \includegraphics[width=.9\linewidth]{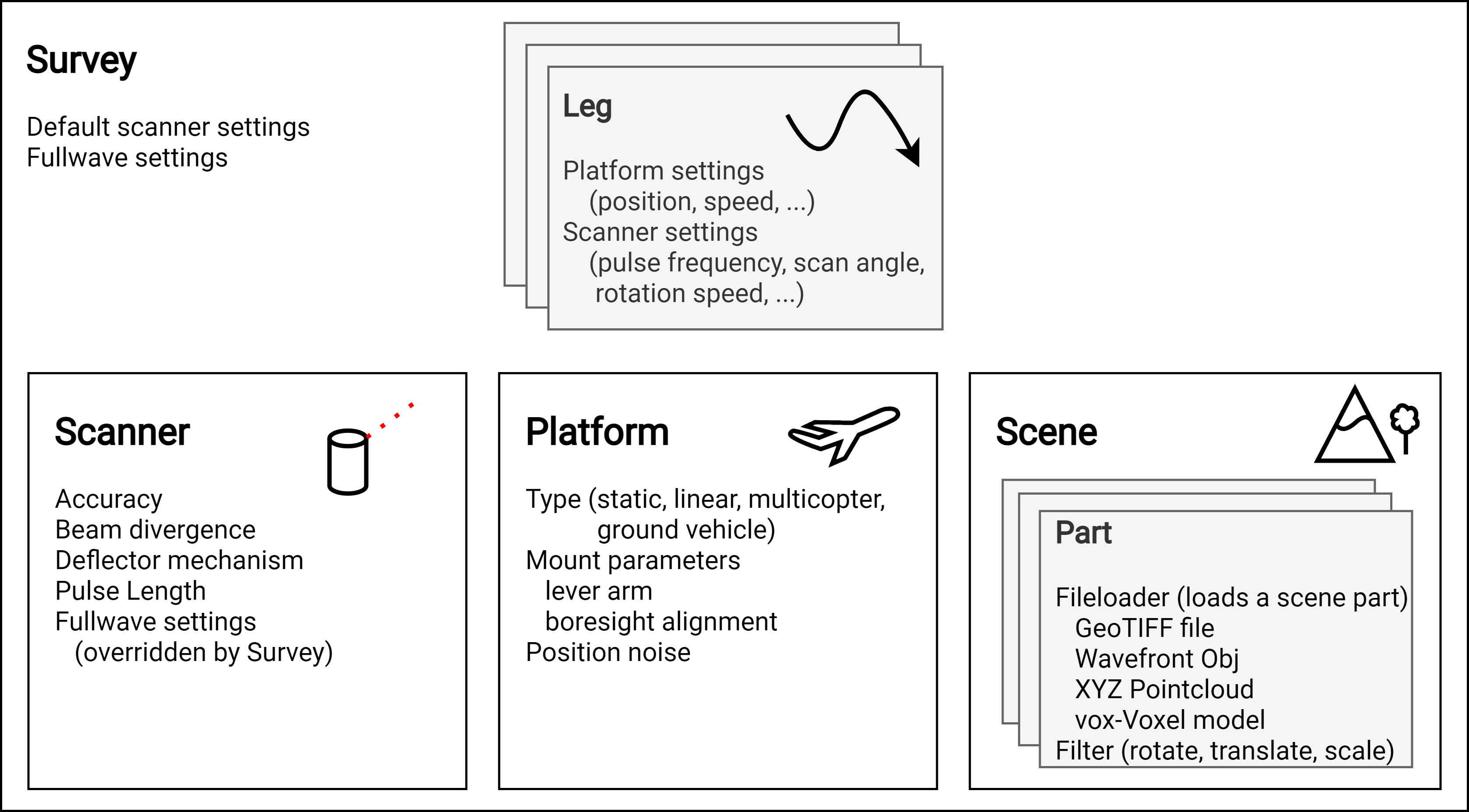}
    \caption{File structure of HELIOS++ survey, scene, platform, and scanner. A survey consists of one or more legs, and a single scanner, platform, and scene, respectively. A scene is built up from one or more parts, which can be of different data type.}
    \label{fig:file_overview}
\end{figure}

The central element in HELIOS++ simulations is a \emph{survey}. A survey contains links to the \emph{scene}, which defines the objects that are scanned, the \emph{platform}, on which the virtual scanner is mounted and moved through the scene, and to the \emph{scanner} itself. Furthermore, a survey contains a number of \emph{legs}, which represent waypoints for the platform. The scene consists of a number of \emph{parts}. Each part represents one input source, for example, a 3D mesh file, or a voxel file. Multiple parts may be combined in a scene, and no limitation to the combination of different data source types is imposed. HELIOS++ uses internationally accepted standard file formats for input and output. These elements are defined through Extensible Markup Language (XML) files, that are referenced using (relative) file paths. XML is a text-based format, which can easily be manipulated using a text editor or an XML editor. Figure~\ref{fig:file_overview} presents the different files along with a subset of the parameters that can be set in the respective files.

A survey may then be run through either i) the command line, where an executable is provided or ii) the Python bindings \verb|pyhelios|. The Python bindings allow access to the simulation parameters as parsed from the XML files, including changing the parameters programmatically for each simulation run. For example, different scanners can be exchanged automatically and simulated in sequence in a single Python script without changing any other input and settings of the simulation. \verb|pyhelios| furthermore allows access to the simulation result, i.e. the point cloud and the platform trajectory, and converts them into a NumPy array either at the end of the simulation run or through a callback function that can be executed every $n$-th cast laser ray. In this way, a live preview of the point cloud acquisition can be implemented in Python to give a visual impression of the ongoing simulation. A Python script distributed with HELIOS++ acts as such a visualiser, and is called with the same commands as the standalone executable. A user may thus quickly switch between using the pure C++ implementation without visualisation or the Python bindings with visualisation, as presented in Listing~\ref{helios_vs_pyhelios}.

\begin{lstlisting}[caption={Comparison between running HELIOS++ as an executable and through the Python wrapper providing an interactive viewer}\label{helios_vs_pyhelios}]
run\helios.exe data\surveys\arbaro_demo.xml --lasOutput --writeWaveform
python pyhelios\helios.py data\surveys\arbaro_demo.xml --lasOutput --writeWaveform
\end{lstlisting}

\begin{figure}[]
    \centering
    \includegraphics[width=.9\linewidth]{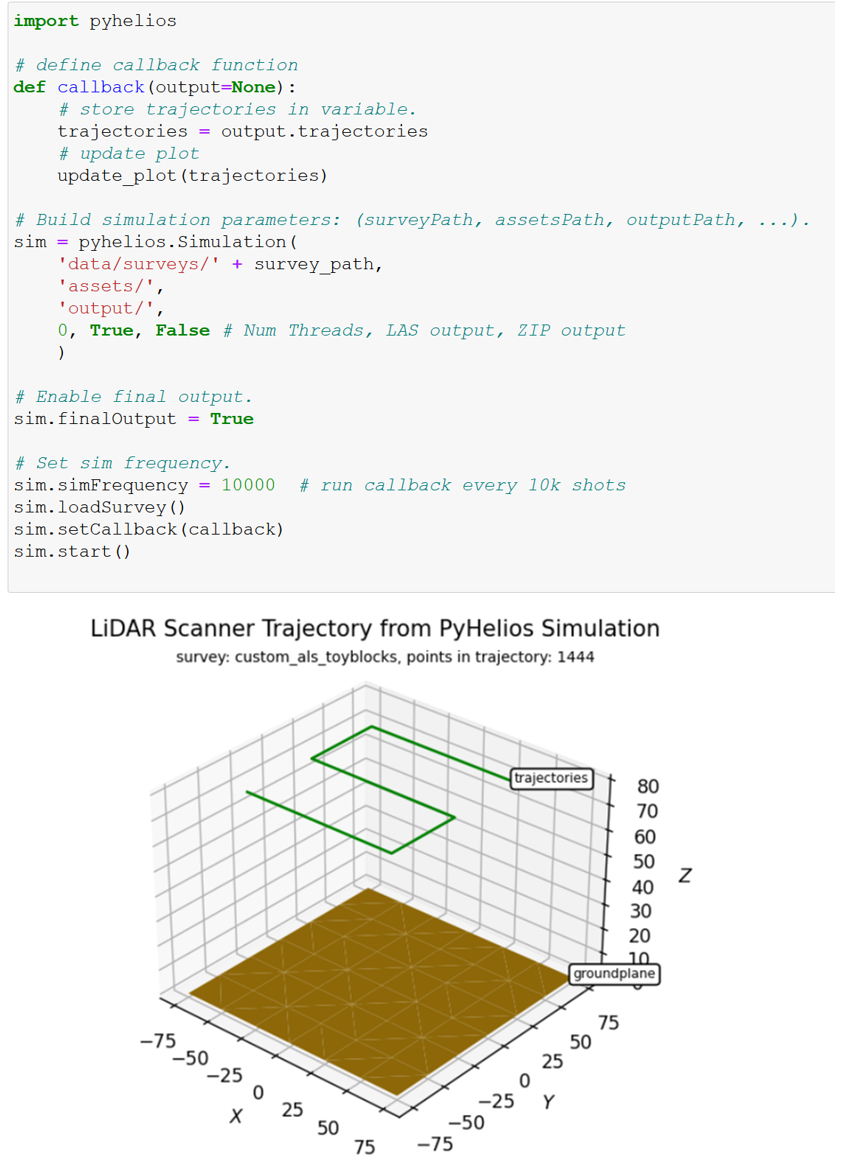}
    \caption{Screenshot of a Jupyter Notebook showcasing the Python bindings of HELIOS++ by plotting the trajectory of a simulation over flat terrain.}
    \label{fig:jupyter}
\end{figure}

With \verb|pyhelios|, it is also possible to combine HELIOS++ with tools like \emph{Jupyter Notebooks}, allowing for explanations along-side code and figures. We include sample notebooks in the documentation of HELIOS++. One of these samples is shown in Figure~\ref{fig:jupyter}.

\subsection{Supported laser scanning platforms}
\label{sec:platforms}
\begin{figure}[h!]
    \centering
    \includegraphics[width=.9\linewidth]{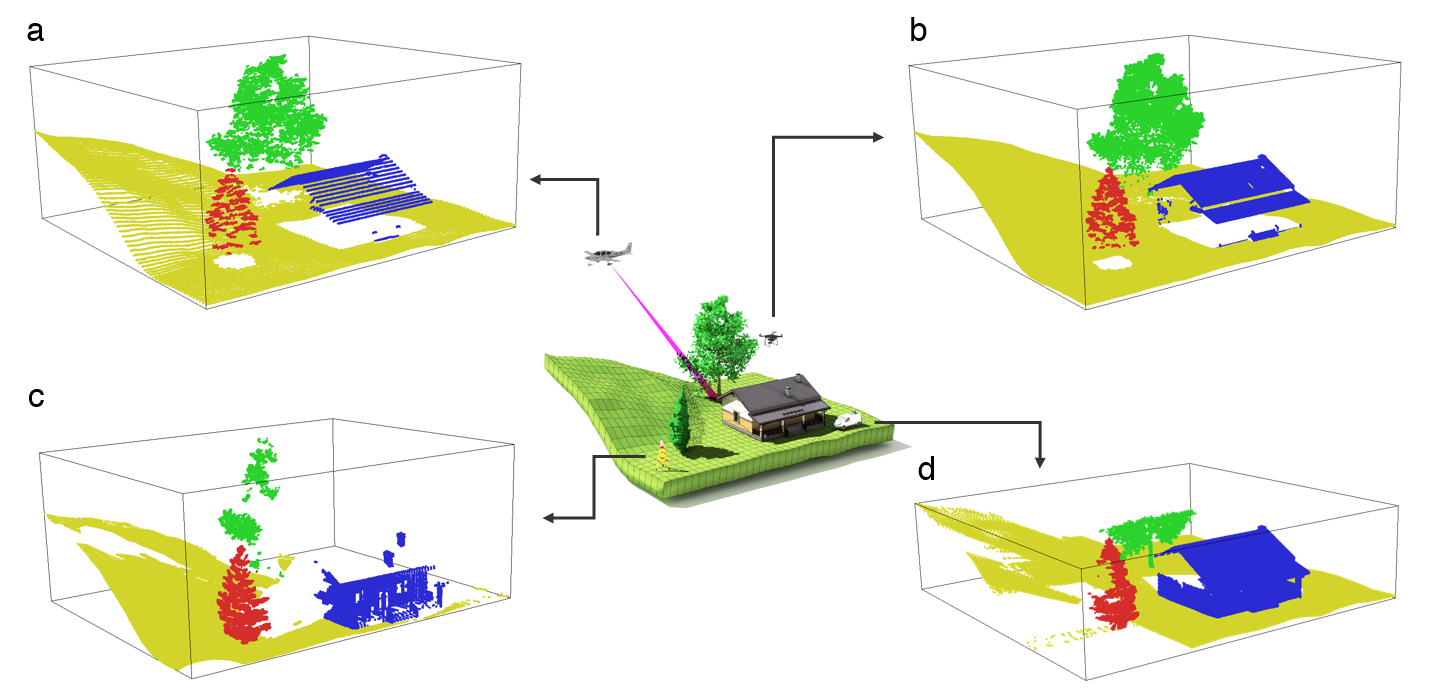}
    \caption{Point clouds resulting from virtual laser scanning of the scene shown in Figure~\ref{fig:geile_figure}, using (a) an airplane, (b) a multicopter, (c) a ground vehicle and (d) a static tripod as platform. Since HELIOS++ records which objects are generating which return, the points can be perfectly assigned to the objects, here illustrated by distinct colouring.}
    \label{fig:geile_results}
\end{figure}

Both static and dynamic platforms are supported by HELIOS++, which can resemble an airplane (\verb|LinearPath| platform), a multicopter (\verb|Multicopter| platform), a ground-based vehicle (\verb|GroundVehicle| platform) or a static tripod (\verb|Static| platform). In the case of the \verb|LinearPath| platform, the vehicle is moved with a constant speed from one waypoint (leg) to the next one. The orientation of the platform is always towards the next waypoint. The \verb|Multicopter| platform additionally simulates acceleration and deceleration of the platform. In the turn mode \emph{smooth}, the platform banks to make more smooth turns at the waypoints instead of stopping and turning on the spot. This mode simulates the \emph{banked angle turns} available in the flight protocols of major drone companies (e.g. DJI). Custom yaw angles for the beginning and the end of each leg can be provided. The \verb|GroundVehicle| platform is bound to elements of the scene defined as ground in the respective material file (cf. Section~\ref{sec:object_models}). Furthermore, it considers maximum turn radii and implements three-point-turns to resemble a car or tractor on which a scanner is mounted (i.e. MLS). The different platforms and scene types (Section~\ref{sec:object_models}) are shown in Figure~\ref{fig:geile_figure}. As expected, choice of platform influences the resulting pointcloud, which is shown in Figure~\ref{fig:geile_results}.

\subsection{Supported laser scanners and scan deflectors}
\label{sec:scanners}
\begin{figure}[h!]
    \centering
    \includegraphics[width=.9\linewidth]{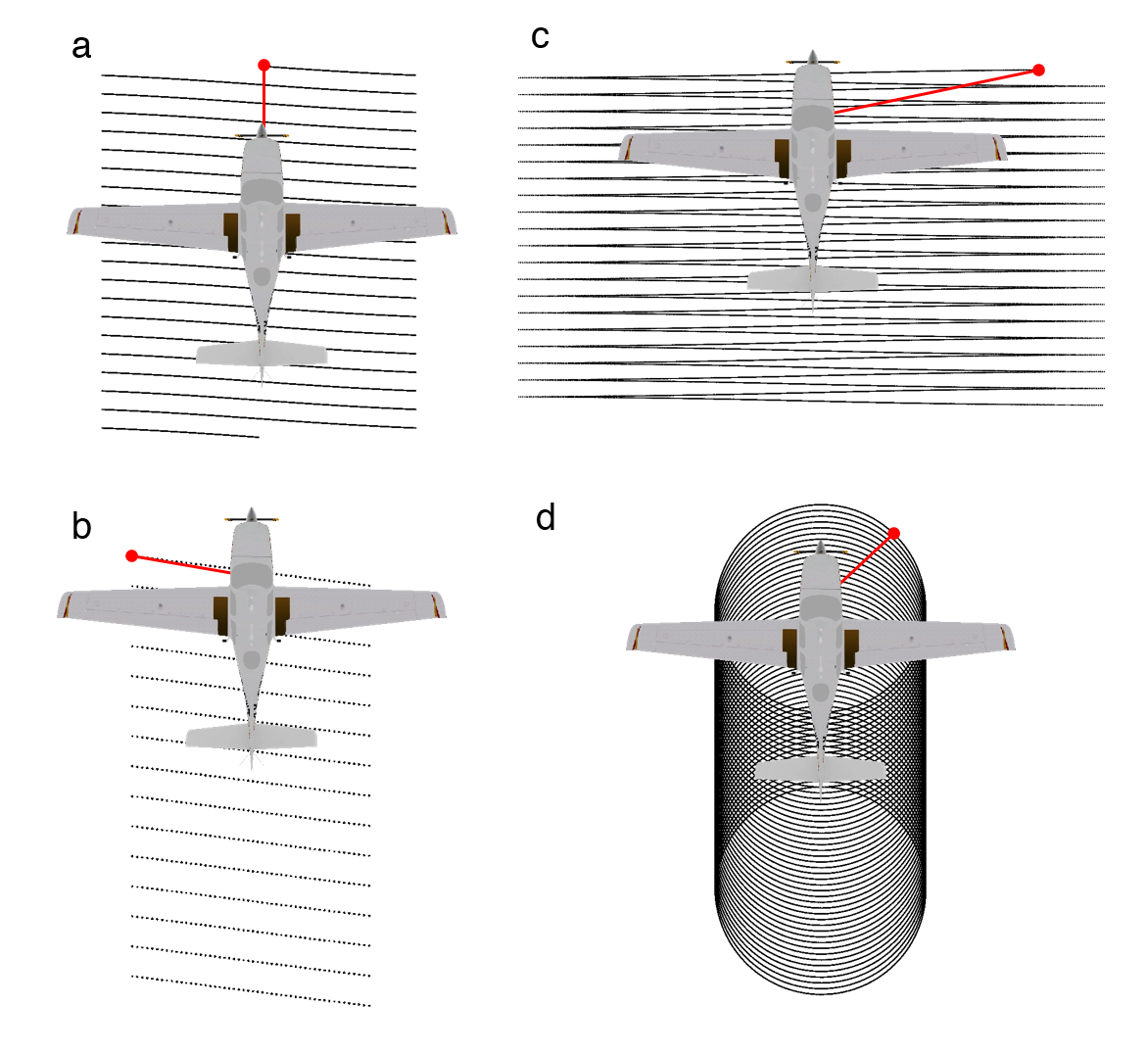}
    \caption{Different scan patterns depending on the deflector used in the simulation: a) rotating mirror, b) fibre-optic line scanner, c) oscillating mirror and d) slanted rotating mirror (Palmer scanner). The patterns shown here result from simulations with HELIOS++ using the respective deflectors, albeit with unrealistic settings of pulse repetition rate and scanning frequency, in order to show the patterns more clearly.}
    \label{fig:scan_patterns}
\end{figure}

The core component of the simulation is the laser scanner, which includes a model for the scan deflector. The choice of the deflector influences the resulting scan pattern (Fig.~\ref{fig:scan_patterns}).

HELIOS++ supports polygonal mirror deflection, resulting in parallel scan lines with even point density (Fig.~\ref{fig:scan_patterns}a), fibre-optics, where the beam is fed through fibreoptic cables to point in different directions, resulting in a similar scan pattern to (a), but without the need of mechanically moving parts (Fig.~\ref{fig:scan_patterns}b), a swinging mirror, which results in a zig-zag-pattern with increased point densities at the extrema (Fig.~\ref{fig:scan_patterns}c) and rotating slanted mirrors, resulting in a conical point pattern at a constant scan angle off-nadir, also referred to as Palmer scanner (Fig.~\ref{fig:scan_patterns}d).

\subsection{Waveform simulation}
\label{sec:fullwave}
During real LiDAR acquisitions, a laser beam sent out from the laser scanner has a finite footprint, i.e. a non-zero area that intersects with the scene. For every infinitesimal point in this area, energy is transmitted back to the detector where it is recorded as the integral of intensities over the area. By considering this received intensity over time, it is possible to extract multiple echoes from one laser pulse, corresponding to multiple targets that were hit by parts of the intersection area, respectively.

\begin{figure}[h!]
    \centering
    \includegraphics[width=.9\linewidth]{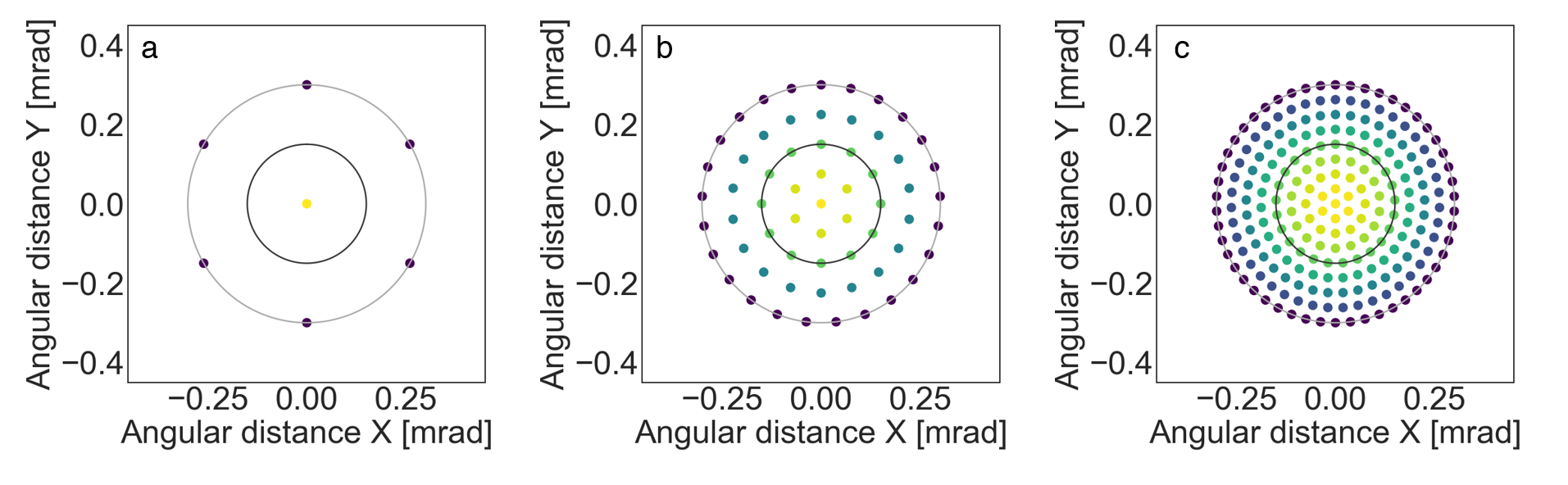}
    \caption{Subray configurations for beam sample qualities of (a) 2 (7 subrays), (b) 5 (93 subrays), and (c) 9 (279 subrays). The colour of the subrays corresponds to the normalised intensity at this location within the beam cone: low (purple) to high (yellow). The black circle represents the single beam divergence (at the $1/e^2$ points, here: \SI{0.3}{\milli \radian}), the gray circle twice the beam divergence.}
    \label{fig:subrays}
\end{figure}

In HELIOS++, the non-zero beam divergence is simulated by subrays, that are sampled in a regular pattern around the central ray (Fig.~\ref{fig:subrays}). Every subray has its own base intensity, which is calculated according to Equation~\ref{eq:beamdiv}, representing a 2D Gaussian power distribution \citep{Carlsson.2001}.

\begin{equation}
    I = I_0 \exp \left(-2r^2/w^2\right)
    \label{eq:beamdiv}
\end{equation}

where $I_0$ [\si{\watt}] is the peak power, $w$ [\si{\metre}] the local beam divergence and $r$ [\si{\metre}] the radial distance from the power maximum, i.e. the ray centre. $w$ is calculated using Equation~\ref{eq:beamdiv_w} from the beam waist radius $w_0$ [\si{\metre}], and the helper values $\omega = \frac{\lambda R}{\pi w_0^2}$ and  $\omega_0 = \frac{\lambda R_0}{\pi w_0^2}$ with $\lambda$ [\si{\nano\metre}] as the wavelength, $R$ [\si{\metre}] the range of the target and $R_0$ [\si{\metre}] the focusing length of the laser.

\begin{equation}
    w = w_0 \sqrt{\omega_0^2 + \omega^2}
    \label{eq:beamdiv_w}
\end{equation}

Every subray is individually cast into the scene and intersected with objects. If it hits an object, a return is generated and recorded by the detector. The respective subray does not continue in the scene through transmission or reflection. In the case of the transmissive voxel model (cf. Section~\ref{sec:object_models}), a subray may either fully traverse a voxel or produce a return, but never both. Therefore, the property of transmissivity of a scene part is coupled to the use of multiple subrays. The power returned from the object is further dependent on the material specified in the scene definition (Section~\ref{sec:object_models}).

The number of subrays generated can be set by the user by providing the \newline \verb|beamSampleQuality| parameter in the XML file of the scanner or the survey. The \verb|beamSampleQuality| corresponds to the number of concentric circles where subrays are sampled. For each circle, the number of subrays is defined as $\lfloor2 \pi i\rfloor$ where $i=1,\dots,$\verb|beamSampleQuality| is the circle index. This ensures that the angular distance between adjacent subrays is approximately constant, i.e. each subray represents a solid angle of equal size. A central subray is always added. Figure~\ref{fig:subrays} shows the subray distribution for three different beam sample qualities.

The pulse shape in time is approximated using bins of a regular, user-defined size via the parameter \verb|binWidth_ns|. Each bin's power is calculated according to Equation~\ref{eq:beam_time}, where $t$ [\si{\nano\second}] is the time and $\tau$ [\si{\nano\second}] is the pulse length of the scanner divided by 1.75 \citep{Carlsson.2001}. $I$ [\si{\watt}] is calculated for each subray according to Equation~\ref{eq:beamdiv}.

\begin{equation}
    P(t) = I \left(\frac{t}{\tau} \right)^2 \exp \left(-\frac{t}{\tau}\right)
    \label{eq:beam_time}
\end{equation}

For every pulse, the subrays are collected as a representation of the full returned waveform, and the recorded power for each bin is taken as the sum of the subray's waveforms, shifted according to the different ranges of the subrays. Range difference is converted to time difference by using the speed of light as a constant ($c=\;$\SI{299792458}{\metre\per\second}). A local maximum filter is then used on the summed waveform to detect peaks, which are regarded as echoes and exported as points. Optionally, a Gaussian may be fitted to the resulting waveform, to get a measure of echo width (i.e. standard deviation of the Gaussian). The position of the point along the range (time) axis, however, is taken from the local maximum, not from the fitted Gaussian, as the outgoing waveform (Eq.~\ref{eq:beam_time}) is not Gaussian.

\subsection{Input scene models}
\label{sec:object_models}
HELIOS++ supports data formats that are (de-facto) standards in the field, and can be created and manipulated with free software such as QGIS\footnote{https://qgis.org/en/site/}, CloudCompare\footnote{https://www.danielgm.net/cc/}, Blender\footnote{https://www.blender.org/} or AmapVOX\footnote{\citet{vincent_2017}, http://amap-dev.cirad.fr/projects/amapvox}.

\subsubsection{Wavefront objects} In wavefront object files (file extension .obj), meshes are represented as lists of triangles (triples of point IDs) and points (triples of coordinates). This allows opaque 3D models of arbitrary complexity. In addition, material properties can be assigned to the objects in so-called material files (file extension .mtl; \citealp{obj}).

\subsubsection{GeoTIFF models} Raster models created with common GIS tools, such as a digital elevation or a digital surface model, can be included. The raster is converted to a triangular mesh on import, where the pixel centres are interpreted as points. Invalid pixels (i.e. no-data values) are ignored in the triangulation, resulting in holes in the mesh. This data type allows an simple representation of the Earth's surface. As prescribed by the file format, only 2.5D-data is supported.

\subsubsection{Point clouds} In ASCII xyz files, point clouds can be used as a raw input to HELIOS++. On import, the point clouds are voxelised using a voxel size defined by the user. In addition, a normal vector used for intensity calculation can be assigned to the voxels (by nearest neighbour or mean reducing) or calculated from the point cloud on the spot (cf. Section~\ref{sec:computational}). Alternatively, all laser rays can be set to have an incidence angle of 0\degree, or the rays are intersected with the actual faces of the cube representing the voxel. Material properties can be defined in the respective scene part definition of the XML file.

\subsubsection{Transmissive voxels} Especially for modelling vegetation, we include support for voxel data containing plant area density information (e.g. created using the AmapVOX software \citep{vincent_2017}, file extension .vox). Multiple modes are supported in this case: 

\begin{itemize}
\setlength\itemsep{0.1em}
    \item opaque voxels, where the voxels are represented by solid cubes with a fixed side length equal to voxel resolution,
    \item adaptive scaling of opaque voxel cubes (with optional random but reproducible shift to avoid regular patterns in the result), where the scaling (side length of a voxel $a$) is dependent on the plant area density $PAD$, the user-defined scaling factor $\alpha$, the base voxel size $a_0$ and the maximum $PAD$, related by Equation~\ref{eq:scaled_voxels}, and 
    \item transmissive voxels (explained in more detail in the following).
\end{itemize}

\begin{figure}[h!]
    \centering
    \includegraphics[width=.9\linewidth]{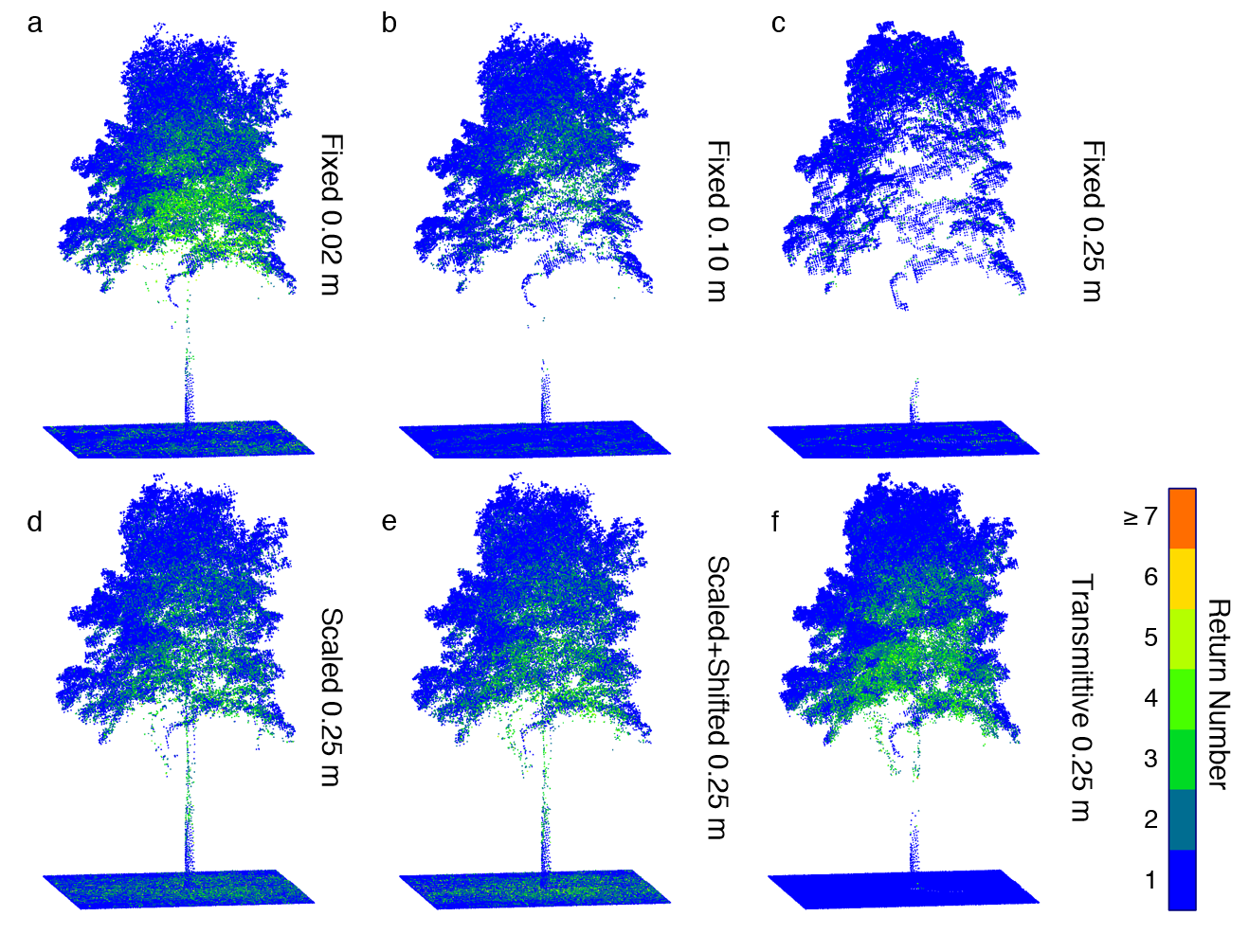}
    \caption{VLS point cloud of a tree based on different voxel modes, using a UAV as a platform. Point clouds are simulated using input voxel models with different fixed sizes (a-c), PAD-dependent scaled voxels (d and e) and transmissive voxels (f).}
    \label{fig:voxel_sim_trees}
\end{figure}

Point clouds resulting from a tree simulation using these different modes are shown in Figure~\ref{fig:voxel_sim_trees}. The opaque voxel models also support the definition of material properties through the scene part's definition. A comprehensive study on the effects of different levels of detail in modelling forests using the first two modes on the extraction of forestry and point cloud parameters is performed by \citet{Weiser2020}.

\begin{equation}
    a = a_0 \left(\frac{PAD}{PAD_{max}}\right)^\alpha
    \label{eq:scaled_voxels}
\end{equation}

Here, $a$ is the side length of the resulting cube, $a_0$ the base voxel size, $PAD$ the plant area density of each individual voxel, $PAD_{max}$ a maximum value for the plant area density and $\alpha$ a user-defined scaling factor.

The transmissive voxels use an extinction approach as presented by \citet{North.1996}. The extinction coefficient $\sigma$ is calculated following Equation~\ref{eq:extinction}:

\begin{equation}
    \sigma = \frac{\mu_L}{2\pi} \int_0^{2\pi} g_L \left| \Omega' \cdot \Omega_L \right| d\Omega_L
    \label{eq:extinction}
\end{equation}

Here, $\mu_L$ is the plant area density (PAD) as defined in the .vox file, $g_L$ is the probability taken from the leaf angle distribution at direction L, and $\left| \Omega' \cdot \Omega_L \right|$ is the cosine of the angle between the incident ray $\Omega'$ and the direction $\Omega_L$. This leaf angle distribution is supplied via a look-up-table as shown in Table~\ref{tab:ladlut}, and examples for planophile, erectophile, plagiophile, extremophile, spherical and uniform distributions are provided.

Subsequently, a random number $R$ is drawn from a uniform distribution $\in [0,1)$. The expectation of the intersection of a subray with a voxel given the extinction coefficient $\sigma$ is then given as in Equation~\ref{eq:intersection_dist}, where $s$ is the distance of an echo after entering the voxel (i.e. traversed distance).

\begin{equation}
    s = \frac{-\ln(R)}{\sigma}
    \label{eq:intersection_dist}
\end{equation}

If the actual length of the subray traversing the voxel is larger than $s$, no echo is recorded for this ray in the voxel, and it is assumed to have transmitted through the voxel, potentially intersecting the next voxel or another object. Otherwise, an echo is created at distance $s$ from where the subray first entered the respective voxel, and the subray is not continued through the scene \citep{North.1996}.

Irrespective of the input type, each scene part can be individually scaled, rotated and translated within the scene, allowing the re-use of models to create multiple instances of similar objects. These transformations can also be manipulated using the Python bindings. For rotations, both extrinsic ("global", fixed coordinate axes) and intrinsic ("local", rotating coordinate axes) modes are supported.

HELIOS++ allows arbitrary combinations of all possible object models in a scene. For example, a GeoTIFF can be used as digital terrain model on which a vehicle is moving, combined with transmissive voxel models of trees, mesh models for tree stems and a point cloud of buildings, which is voxelised by HELIOS++ (Figure~\ref{fig:geile_figure}). Such combination allows highly versatile use of the scene definition, while considering shadowing and overlapping effects between the different types of object models, and is one of the key advantages over other VLS software.

\subsection{Output format and options}
\label{sec:output}
The simulated point clouds can be written either as LAS file (Version 1.0, point format 1), according to the format definition by the American Society of Photogrammetry and Remote Sensing \citep{asprs2011}, as LAZ file, which is a lossless compression of LAS provided by the LASzip library\footnote{https://laszip.org/, \citet{isenburg2013laszip}}, or as ASCII file, where the point coordinates and attributes are written in columns. The LAS/LAZ formats allow for smaller file sizes and faster read/write access (cf. Section~\ref{sec:performance}), while the ASCII format can be parsed by any program working with text files.

In addition to creating compressed LAS files, HELIOS++ can also compress output ASCII files, if smaller output sizes are required. Uncompressing LAZ or compressed ASCII files is possible by invoking HELIOS++ using the \verb|--unzip| flag.

In the simulation of full waveforms (see Section~\ref{sec:fullwave}), the echo width is optionally estimated for every recorded return. Since this estimation uses a non-linear least squares method and is hence expensive to calculate, the estimation has to be switched on explicitly by the user. In typical use cases, this increases the runtime of the simulation by a factor of 1.3.

If the full waveform is to be exported, a separate ASCII file will be written. This file contains information about every beam that generated at least one hit consisting in the beam origin and the beam direction, as well as the sampled returned waveform. The bin size as well as the maximum length of this sampled waveform can be defined by the user (Section~\ref{sec:fullwave}). To connect the point cloud output with the full waveform output, the point cloud has an attribute \verb|fullwaveIndex|, which corresponds to the \verb|fullwaveIndex| in the ASCII waveform output. Multiple points, if coming from the same beam, may have the same \verb|fullwaveIndex|. The output created by HELIOS++ can be easily converted to standard waveform formats such as PulseWaves and LAS 1.4 WDP.

Furthermore, HELIOS++ comes with an option to export the trajectory of the platform. The time interval between successive platform positions can be defined by the user in the survey XML file. In addition to the time and the position of the platform, the attitude angles of roll, pitch and yaw are exported.

\subsection{Randomness and repeatability}
\label{sec:randomness}
The trajectory of the platform, the scanner definitions and the ray-scene interactions with the exception of transmissive voxels are deterministic. To allow for more realistic point clouds, random noise sources may be introduced at various points of the simulation. A single distance measurement is attributed with a random ranging error, drawn from a normal distribution with parameters defined for each scanner. Additionally, random platform noise can be added to simulate trajectory estimation or scan position localisation errors. By default, the system time is used to generate randomness, which results in different outcomes for every simulation run.
However, the ability to produce the identical results in repeated simulation runs is a major advantage of VLS over real data acquisitions and may be aspired for certain use cases. To allow for repeatable survey results while using randomness, there are additional options to define a custom seed for pseudo-randomness generation. Still, when using multithreading, each thread accesses the randomness generator in a non-deterministic order, resulting in different outputs for every simulation run. This can be avoided by running HELIOS++ in single-threaded mode and with a fixed seed. This will generate the exact same result in repeated runs, even for transmissive voxels, as the random number drawn from the uniform distribution is also created based on the seed.

\section{Applications of laser scanning simulation}
\label{sec:applications}
To investigate the usability of the HELIOS++ concept of providing a sensible trade-off between physical reality and computational complexity, along with the support of generic input files to create the 3D scene, we present applications of laser scanning simulation. This literature survey is based on studies that use HELIOS++'s predecessor, HELIOS \citep{Bechtold_2016} or pre-release versions of HELIOS++. Since the functionality of HELIOS, apart from the visualisation module, is fully integrated in HELIOS++, the publications presented in the following provide an adequate review of simulation applications.

The analysis is conducted on all publications that cite the original publication of \citet{Bechtold_2016}, and that actively use HELIOS in their research according to the published article. It consists of journal papers, conference contributions and posters. The following questions are posed to analyse the contents of each publication.

\begin{itemize}
\setlength\itemsep{0.1em}
    \item What is the scientific target of the simulation?
    \item How many simulations are carried out on which platform (ALS, TLS, ULS, MLS)?
    \item Which type of model is employed to compose the scene for the simulation?
    \item Which parameters are extracted from the point cloud, if applicable?
    \item What is the resulting VLS point cloud compared to?
\end{itemize}

The synthesis of the literature review allows a grouping of the publications and of the purposes of laser scanning simulation into four main categories: (1) optimising or analysing different scan settings and acquisition modes, (2) comparing parameters extracted from simulated data to error-free ground truth, (3) generating training data for supervised machine learning, and (4) testing and development of novel or future algorithms and sensors.

\subsection{Data acquisition planning and scan setting effects}
Data acquisition planning at its core is an optimisation problem. The goal is to acquire data that is fit for the purpose of the analysis by minimal effort. This can be a minimum number of scan positions, flight lines, etc., that is sufficient for the specific requirement to the data, such as coverage of the scene regarding occlusion or a certain point density. While flight planning tools and visibility analysis-based methods can be used to create potential acquisition plans, they usually lack methods to verify the fitness for use of obtained data.

With virtual laser scanning, assuming that at least a coarse model of the area of interest is available, a 3D point cloud can be generated and tested for its fitness directly in the planned application. There is no need to define proxy metrics like target point density, required accuracy and overlap as required by simple survey planning tools. This way, users can ensure that the acquired data will meet all requirements such as coverage, adequate representation of geometry, and resolution before actually going to the field to collect the data, by running their analyses on the simulated point clouds and interpreting the results. Similarly, the effects of different scan settings on parameters extracted from the simulated point clouds can be studied with HELIOS++, as single variables can easily be manipulated in a way that is isolated from all other influences, including environmental influences, which are very difficult to control in repeated real-world acquisitions. For example, the effects of flying height and maximum scan angle on the resulting ground point density and resolution (i.e. illuminated area per beam) may be analysed. Similarly, TLS scans of different resolution can be simulated and the effect of resolution on the results of data analysis (e.g. the quality of extraction of tree stems) can be quantified. 

While the quality of the simulation in terms of being physically realistic highly depends on the input models, even a coarse model can be useful to estimate occlusion and resulting point densities. Such analyses, carried out prior to real data acquisitions, can save valuable time in the field. Using a Monte-Carlo approach, multiple simulations using different parameters are carried out to create the optimal (in terms of number of positions, time, etc.) acquisition plan.

Existing publications in this category include \citet{Backes_2020}, who use a digital surface model created from photogrammetry to simulate acquisition of an alpine valley by TLS and ULS. They estimate the minimal detectable change, to optimise scan positions and trajectories for change analysis. Similar analyses, but with focus on resulting point density and completeness of data acquisition, are carried out based on a DEM provided by public agencies by \citet{Lin_2019}. They are able to show that a downsampled scene model can provide accurate measures for simulated point densities when also reducing the pulse frequency. This validates the simulation's representation of spatial scales.

A validation of scan position planning based on viewshed analysis is carried out with respect to achieved accuracy, point density and completeness by \citet{Previtali_2019}. Their use cases are acquisitions of complex archaeological sites, where complete coverage is often difficult to achieve due to occlusions. In two examples, they optimise the positions of 97 and 16 scan positions, respectively, to scan a basilica and part of an ancient food storage in Italy.

The effect of scan settings on point cloud-based parameters is analysed for forestry settings by \citet{Hammerle_2017}, who extract understory tree heights from TLS and ULS data. They use 3D models created with the Arbaro tree generator\footnote{http://arbaro.sourceforge.net/ \citep{Weber.1995}}, on which they densely sample points to create a reference point cloud as ground truth. They find a favourable trade-off between acquisition effort and accuracy of results for a number of three TLS scan positions around a tree object of interest \citep{Hammerle_2017}. Similarly, \citet{Li_Mu_Soma_Wan_Qi_Hu_Zhang_Tong_Yan_2020} create 3D tree models using ONYXTREE\footnote{http://www.onyxtree.com/} and evaluate the influence of scan parameters on extracted values of diameter at breast height (DBH), tree height, stem curve and crown volume. They succeed in reducing the root-mean-squared error on these values by iteratively adapting scan parameters. 

\citet{Weiser2020} analyse the different opaque voxel models at different scales and their effect on common tree metrics, showing that a scaled voxel model requires much less complexity (i.e., allows for a larger voxel size) than binary voxels at a fixed scale. \citet{Schaefer_2020} use the novel transmissive voxel-support of HELIOS++ for similar analyses with ALS and ULS data.

\subsection{Algorithm and method evaluation: validation and calibration}
In numerous point cloud analysis methods, the objective is to extract certain parameters describing the objects of interest. For example, in forestry applications, ALS and TLS point clouds are commonly used to derive the diameter at breast height of trees, crown radii, tree heights, and tree species \citep{giannetti_2018}. To calibrate the extraction algorithms as well as to validate the results, in-situ measurements are required, which are laborious and costly to acquire, and not free of error.

An alternative to this in-situ data acquisition can be provided by {HELIOS++}, if the parameters to be extracted from the point cloud can be derived from the objects within the scene, or the scene is created dynamically according to the parameters. For example, a method may be designed to measure DBH from a point cloud. In VLS, the true values of DBH can be derived from the stem models, or the virtual tree models themselves may be generated according to given values of DBH.  In addition to being error-free, the domain of the parameters (e.g. the range of DBH values) used in the simulation can be defined by the user. A simulated example can therefore be picked to have exactly the properties needed in a specific application (say, use case-specific DBH values between \SIrange{20}{25}{\centi\metre}), whereas real objects with the required properties as ground truth may be hard to find.

In addition to parameter extraction, perfect ground truth can be used to evaluate the performance of classification algorithms. Currently, classification methods are difficult to compare, as authors use different evaluation approaches, mainly due to the lack of semantic reference data. Since in VLS the interaction between the laser beam and the scene can be attributed to the exact mesh face that is hit during ray casting, and therefore to a specific object, the ground truth data is free of error even in highly complex and multi-echo scenarios. This has compelling advantages:  First, reference data can be created automatically, drastically cutting costs and providing the possibility of generating massive amounts. And second, it removes labelling errors, as manual labelling will always include some degree of human error.

Considerable research using HELIOS to validate or calibrate extracted parameters has been undertaken in multiple forestry applications. \citet{Liu_Skidmore_Wang_Zhu_Premier_Heurich_Beudert_Jones_2019} and \citet{Liu_Wang_Skidmore_Jones_Heurich_Beudert_Premier_2019} estimated leaf angles on trees, where ground truth is practically impossible to obtain, because in real settings, leaves are continuously moving due to wind. VLS allows to validate their approach of leaf angle estimation and subsequently apply it to real data. \citet{Wang_Schraik_Hovi_Rautiainen_2020} validate their calculation of photon recollision probability over spatial locations using VLS data. \citet{Zhu_Liu_Skidmore_Premier_Heurich_2020} use two different methods to assess Leaf Area Index (LAI) and compare these methods with ground truth obtained from the object models, allowing them a comparison with the true LAI of the input objects. For tree segmentation, ground truth is also difficult to obtain, as tree canopies often intersect each other. By using HELIOS, \citet{Wang_2020} and \citet{Xiao_Zaforemska_Smigaj_Wang_Gaulton_2019} obtain perfect ground truth for training and validation of their segmentation methods. \citet{Wang_2020} achieve 2.9~\% and 19.8~\% RMSE for tree height and crown diameter estimates.

In a non-forestry application, road curves are reconstructed and the reconstruction is compared with the model parameters \citep{Zhang_Li_Guo_Yang_Wang_2019}, achieving a relative accuracy of 0.6~\% in circle radii estimation using VLS data. Requirements for Building Information Modelling (BIM) are evaluated by \citep{Rebolj2017}, who generate around 100 point clouds to ascertain the influence of parameter values. They define accuracy criteria for the successful identification of building elements in the scans. \citet{Bechtold_2016b} test a segmentation tool for rock outcrops by using HELIOS as a simulator. They show the value of simulated test data for method development by easily generating point clouds with different occlusions and scan settings, resulting in a multitude of point densities and point patterns.

\subsection{Method training}
With the advent of neural networks as supervised machine learning method in geospatial domains, the need for training data has grown almost indefinitely. While raster-based approaches can make use of existing pre-trained networks by domain transfer \citep{Pires_de_Lima_2019}, no such networks exist for point-based deep learning such as PointNet/PointNet++ \citep{qi2017pointnet++}.

Simulated data, though only replicating parts of reality, can be used by neural networks to learn basic descriptors which describe point cloud neighbourhoods, e.g. planes, corners or edges. From these descriptors, higher-level features are derived, which are subsequently used in classification or regression \citep{Winiwarter2019}. 

Once a network has learned to represent data in form of these features, it can be adjusted to real data by adding relatively small amounts of training data, in approaches shown to work for the image domain \citep{Danielczuk2019}. Further research is needed on this approach especially for point cloud data, but HELIOS++ allows easy and fast generation of labelled training data, which does not suffer from ground truth errors.

As an example application of this purpose, \citet{Martinez_2019} use HELIOS-simulated data to train and evaluate a semantic classification of an urban scene created from OpenStreetMap\footnote{https://www.openstreetmap.org/} models. The use of VLS allowed a quantification of their classifier's total error, which amounts to 0.5~\% on simulated point clouds.

\subsection{Sensing experimentation}
The fourth category summarises publications concerned with the development of novel sensors and methods, such as \citet{Park_2020}, who present a new Time-of-Flight sensor and compare its results to a HELIOS simulation. 
In general, the parameters of the virtual sensors can be tuned to resemble a non-existent sensor, the performance of which can then be simulated without the need of actually building a prototype. Especially when looking for potential improvements of current sensors, this enables the identification of weak links or bottlenecks for certain use cases.
 
More in-depth experimentation is also conceivable, where e.g. a novel deflector model shall be simulated. Due to the open-source license, a developer may take the HELIOS++ framework and would only have to implement a new deflection method, whereas the scene and all other components can be used as is to run a simulation, testing the usability of the novel deflector model.

Especially when considering the short lifecycle of current hardware, simulation may be the only way to ensure fitness for use of the sensor. Equivalent simulators are widely used in remote sensing, as especially tools working with data acquired by satellites can be developed using the simulated data, and are then ready to use as soon as the first real data is delivered.

\section{Computational considerations}
\label{sec:computational}
Since modern laser scanners can measure millions of points per second, it is crucial to have an efficient implementation for the ray-scene intersection. In this section, we first present theoretical considerations on the ray tracing implemented in HELIOS++, the modelling of vegetation and options of generating large and complex scenes. In Section~\ref{sec:performance}, a comparison between HELIOS++ and its predecessor HELIOS is carried out.

\subsection{Ray tracing implementation using a kD-Tree}

A scene $S$ in HELIOS++ context can be mathematically described as a set of primitives, so $S = \{P_{1}, \hdots ,P_{n}\}$. For each primitive $P$ its boundaries are defined considering an axis-aligned bounding box, its centroid, and the set of vertices $V = \{v_{1}, \hdots, v_{m}\}$ composing the primitive. Each primitive supports rotation, scaling and translation together with a material specification defining its reflectance and specularity. Certain primitives, such as transmissive voxels, also support a look-up table which can be used for vegetation modelling, as explained in Section~\ref{sec:vegmodel}.

The scene building process consists in generating the set of primitives composing the scene. Multiple input sources are supported (cf. Section~\ref{sec:object_models}), so it is possible to build a scene considering Wavefront Object files, point clouds as ASCII files specifying $(X, Y, Z)$ coordinates for each point, GeoTIFF files, or a custom voxel file format based on AMAPVox \citep{vincent_2017}. When building a scene, different sources can be considered, as each one is associated to its own scene part. It is also possible to apply aforementioned affine transformations to an entire scene part. This can be used, for instance, to load the same object multiple times in one scene and placing it in different locations through translations. Saving and loading already built scenes is relying on boost serialisation technology~\citep{boostserial}.

Ray intersections are computed through a recursive search performed over a kD-Tree containing all primitives~\citep{Bentley.1975}. Let $O$ be the ray origin and $\hat{v}$ the normalised ray direction vector. When recursively searching through the kD-Tree starting at $O$, $\hat{v}$ is used to consider which node must be visited until a leaf node is reached. For each recursive search operation $s$, coordinates are analysed, so $s \equiv 0 \mod 3$ means the $X$ coordinate splits the space, $s \equiv 1 \mod 3$ means the Y coordinate splits the space, and $s \equiv 2 \mod 3$ means the Z coordinate splits the space. Once inside a leaf node, ray intersections with respect to each primitive are computed. The minimum distance intersection $t_0$ is the time the ray needed to enter the primitive while $t_1$ is the time the ray needed to leave it. It is possible to have only $t_0$ determined, as is the case for triangles, since they are only intersected once per ray.

For the special case of primitives which support multiple ray intersections, such as transmissive voxels, the process is repeated considering consecutive origins. Suppose we have a transmissive voxel intersected by a ray $\{O_{1}, \hat{v}\}$: If this voxel lets the ray pass through it, then the next ray intersection will be found considering the ray $\{O_{2}, \hat{v}\}$, with $O_{2} = O_{1} + (t_{1}+\varepsilon) \hat{v}$, where $\varepsilon$ is a small decimal number to assure getting out of the previously intersected primitive.

\subsection{Vegetation modelling: Transmissive voxels}\label{sec:vegmodel}

\begin{table}[]
\centering
\begin{tabular}{ll|l}
\textbf{Beam horizontal component} & \textbf{Beam vertical component} & \textbf{Hit probability $g_L$} \\ \hline
1.000000                           & 0.000000                         & 0.424413                       \\
0.999683                           & 0.025180                         & 0.424682                       \\
0.998732                           & 0.050345                         & 0.425489                       \\
$\vdots$                           & $\vdots$                         & $\vdots$                       \\
0.009444                           & 0.999955                         & 0.848789                       \\
0.000000                           & 1.000000                         & 0.848822                      
\end{tabular}
\caption{Values from a look-up table (LUT) for the hit probability $g_L$ at a given beam direction $L$, represented by horizontal and vertical component. The numbers here correspond to an erectophile distribution and are obtained using a numerical integration method on the formulae from \citet{North.1996}. The probability $g_L$ is normalised over $2\pi$.}
\label{tab:ladlut}
\end{table}

One of the new functionalities of HELIOS++ is vegetation modelling through transmissive voxel primitives. For this purpose, a look-up table for leaf angle distribution is used to compute ray intensity with respect to a $\sigma$ (cross-section) value obtained from it. The values in this look-up table represent a hit probability $g_L$ given the horizontal and vertical component of the beam vector  (Tab.~\ref{tab:ladlut}). The intensity $I$ is calculated according to Equation~\ref{eq:ladlut_intensity} where $P$ is the emitted power from Equation~\ref{eq:beam_time}~\citep{Carlsson.2001}, $d$ is the distance between ray origin and intersection point, $\alpha^2$ is the square of the scanner receiver diameter, $\beta^2$ is the square of the scanner beam divergence and $\lambda$ is the product between atmospheric factor and scanner efficiency. The $\sigma$ value can be seen as function of ray direction vector $\sigma(\hat{v})$, so it will have a different value depending on ray incidence.

\begin{equation}
    I \propto \lambda\sigma \frac{P\alpha^2}{4{\pi}d^{4}\beta^2}
\label{eq:ladlut_intensity}
\end{equation}

Voxels can operate in transmissive mode. This implies the voxel will let the ray continue and not return a signal after intersection if $\sigma = 0$. If $\sigma > 0$, then the return of the ray is randomly determined by sampling a value $u$ from a uniform distribution $\in[0,1)$ and computing $s = \frac{-\log(u)}{\sigma}$. If the value of $s$ is greater than the distance between both intersection points at the voxel, the ray will continue. Otherwise, the ray will stop at the voxel. For the non-transmissive mode, only voxels with transmittance $1.0$ will allow rays to continue.

Thus, by using detailed voxels operating in transmissive mode together with an appropriate look-up table specification, HELIOS++ is capable of simulating laser scanning of vegetation from a precomputed leaf angle distribution following \citet{North.1996}. To achieve high-quality output, it is recommended to apply individual leaf angle distributions for each vegetation type within a scene. 

\subsection{Strategy to handle large and complex scenes}

When loading point clouds from ASCII xyz files, big files might not be entirely containable in memory. For this purpose, a two-stage algorithm is used to digest point clouds of arbitrary size. The first stage simply finds the minimum and maximum values for each coordinate and counts the total number of points. The second stage builds all necessary voxels to represent the point cloud inside HELIOS++. For this purpose, a voxel grid is allocated. Then, voxels which have points inside them are built as HELIOS++ primitives. For each voxel its spatial coordinates $(X,Y,Z)$, normal vector components $(N_x, N_y, N_z)$ and colour components $(R,G,B)$ are considered, if available, as may be the case for photogrammetric point clouds.

The colour for the voxel is computed as the average of each colour component for all points inside the voxel. We approximate the sRGB color space by averaging the squares of the red, green, and blue values, and taking the square root of this average as the resulting value for the voxel. If normals are provided, the voxel normal can be determined either as the normal of the point which is closest to the voxel centre or the average of each normal component for all points.

In case no normal vectors are provided, HELIOS++ can estimate them using singular value decomposition (SVD, \cite{Golub.1965}). All points inside a voxel are considered to build a matrix of coordinates, for which singular values and singular vectors are obtained. Then, the singular vector of the smallest singular value is the orthonormal vector defining the plane which best fits the point set in terms of the smallest sum of squared orthogonal residuals. It can hence be understood as the voxel normal vector. The normal estimation method can be applied to the entire input point cloud in a single stage for small point clouds. If the size of input point cloud is too big with respect to RAM, the normal estimation is performed in batch mode, dividing the workload into smaller parts. Each batch extracts points inside its voxels while ignoring points outside its scope.

\subsection{Performance comparison}
\label{sec:performance}
As the direct successor of HELIOS \citep{Bechtold_2016}, it is interesting to compare the performance of HELIOS++ with the original implementation in Java. However, HELIOS++ was not just a port of the code, but also comes with multiple computational improvements over its predecessor. One of these improvements concerns the ability for the user to better leverage accuracy at the cost of runtime, or vice versa, by setting parameters accordingly. For some tasks, a rough, thereby faster simulation may suffice, whereas for other tasks very detailed simulation is required, but smaller sample sizes can be used or long processing times are acceptable.

Another improvement concerns the binning mechanism for the full-waveform simulation, which is used for maxima detection even if the waveform is not written to an output file. In HELIOS++, we use the parameters \hbox{\verb|binSize_ns|} and \verb|maxFullwaveRange_ns| for this binning, where the bin size is used for both the outgoing pulse and the returned waveform. To limit the impact on performance of very low-incidence rays with a high sampling quality, the user can provide a maximum length of the recorded waveform, beyond which any further echoes are discarded.

To compare the performances of HELIOS++ (Version 1.0.0) and HELIOS (Version 2018-09-24), we carry out a number of simulations using different parameter settings, and record the runtime as well as peak memory usage. We present three different scenes, at three different complexity levels, and make use of the option in HELIOS++ to write different file types and to skip the echo width determination, if not required.

\begin{figure}[h!]
    \centering
    \includegraphics[width=.9\linewidth]{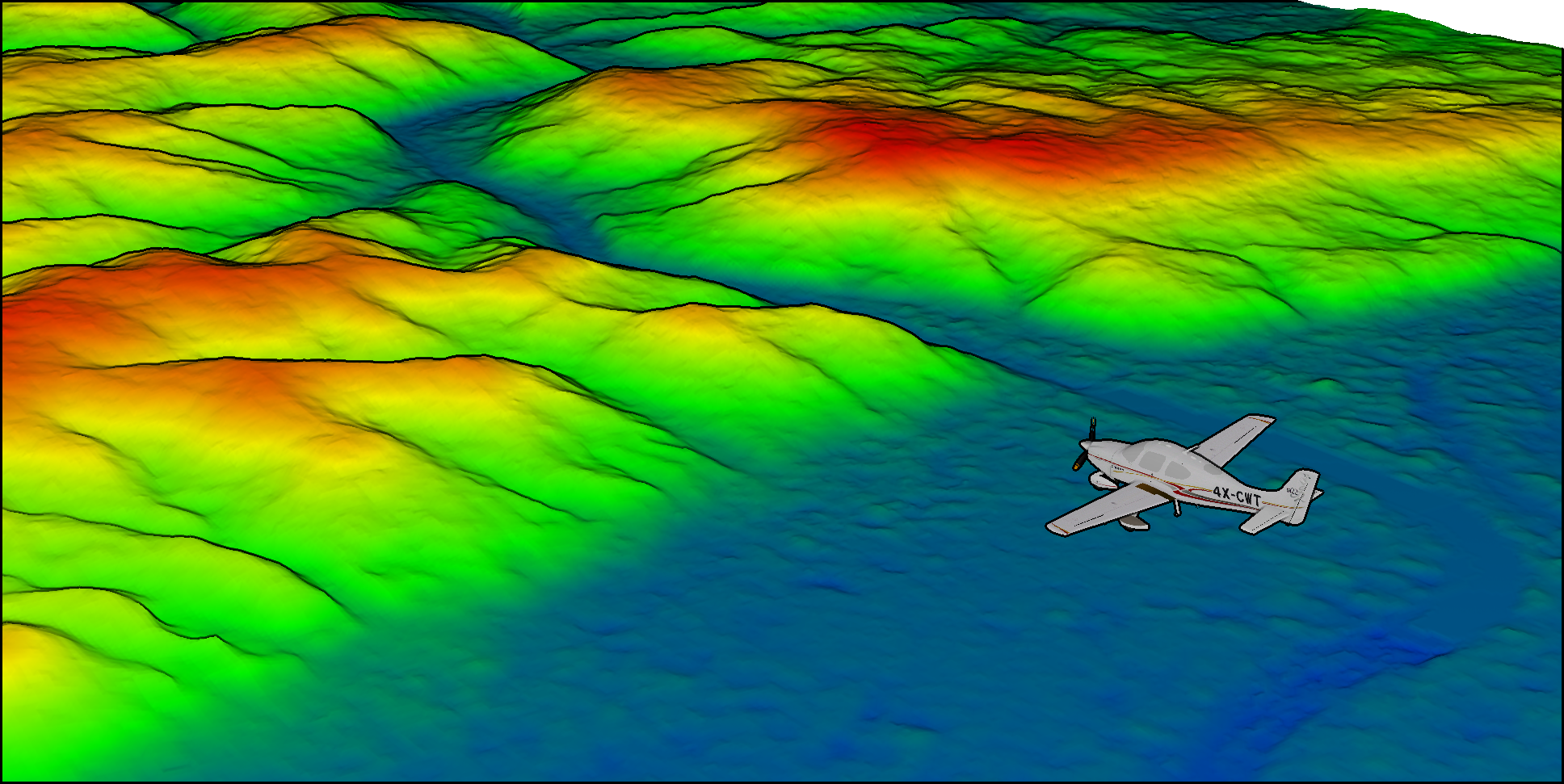}
    \caption{Visualisation of example survey for airborne laser scanning (ALS), using a digital terrain model as object and a standard ALS instrument as scanner. The 3"-SRTM model by the USGS is used as input raster.}
    \label{fig:als_demo}
\end{figure}

The first example scenario represents an ALS survey over terrain, which is created by loading a digital terrain model in GeoTIFF format as described in Section~\ref{sec:object_models}. Since the GeoTIFF loader in the Java version was faulty and this version is no longer maintained, we created a mesh in Wavefront Object format as an input for the Java version. We simulate two flight lines at an altitude of \SI{1500}{\metre\, asl} over terrain with an extent of \SI{26x17.8}{\kilo \metre}. As scanner we use a \emph{Leica ALS50-II}, and a Cessna SR-22 as platform. The scene is shown in Figure~\ref{fig:als_demo}.

\begin{figure}[h!]
    \centering
    \includegraphics[width=.9\linewidth]{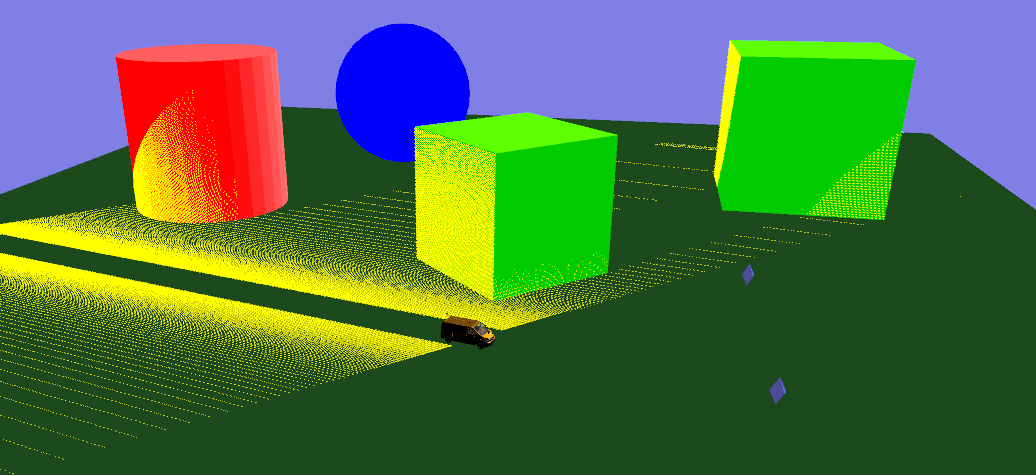}
    \caption{Visualisation of example survey for mobile laser scanning (MLS), with a car driving between objects of simple geometry. The yellow points show the acquired point cloud of scanned parts.}
    \label{fig:mls_demo}
\end{figure}

The second scenario represents a mobile laser scan of an urban area, where a car is driving through downtown buildings modelled as prisms, cylinders and pyramids (Figure~\ref{fig:mls_demo}). The trajectory of the car is \SI{208}{\metre} long and its average speed is \SI{20}{\metre \per \second}. The scanner is an oblique-mounted \emph{RIEGL VUX-1UAV}.

\begin{figure}[h!]
    \centering
    \includegraphics[width=.9\linewidth]{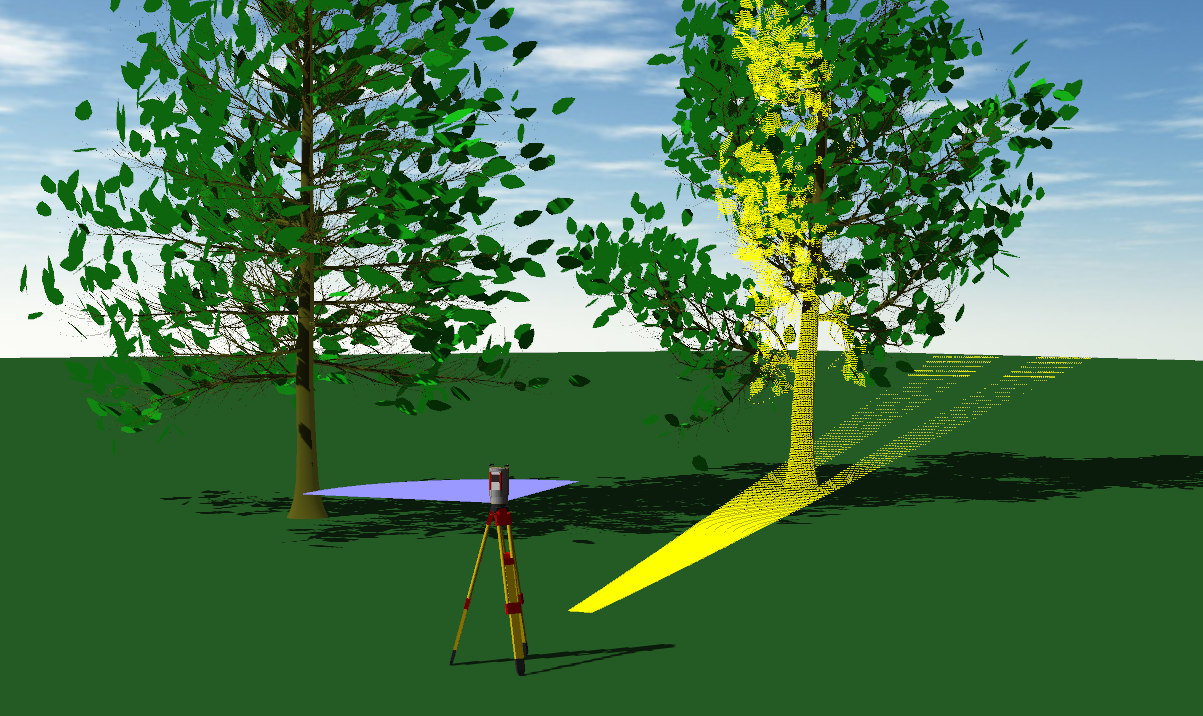}
    \caption{Visualisation of example survey for terrestrial laser scanning (TLS), scanning two highly complex tree models created using the Arbaro Tree Simulator. The yellow points show the acquired point cloud of scanned parts.}
    \label{fig:tls_demo}
\end{figure}

In the third and final scenario, we present a TLS survey of a vegetation scene with a large potential for multi-echoes, as it represents two trees generated with the Arbaro Tree Simulator \citep{Weber.1995}, scanned using a \emph{RIEGL VZ-400} from two static positions. This simulation has a large number of geometric primitives. A visual is depicted in Figure~\ref{fig:tls_demo}.

Each simulation scenario was carried out three times on an Intel-i9-7900 @ \SI{3.3}{\giga\hertz} with \SI{64}{\giga\byte} of RAM, and I/O on an SSD connected via SATA. The average runtime and memory consumption values are given in Table~\ref{tab:performance}.

\begin{sidewaystable}[]
\centering
\begin{tabular}{l|l|lll}
                                                                                           & \begin{tabular}[c]{@{}l@{}}\textbf{HELIOS (Java)} \\ Version 2018-09-24 \end{tabular}                                                                  & \multicolumn{3}{c}{\begin{tabular}[c]{@{}l@{}}\textbf{HELIOS++} \\ Version 1.0.0 \end{tabular} }                                                                                                                                                                                                                                                      \\
Echo width                                                                                 & \checkmark                                                                                    & \checkmark                                                                                   &                                                                                              &                                                                                               \\
Waveform output                                                                            & \checkmark                                                                                    & \checkmark                                                                                   & \checkmark                                                                                   &                                                                                               \\
File format                                                                                & XYZ                                                                                           & XYZ                                                                                          & XYZ                                                                                          & LAS                                                                                           \\ \hline
\textbf{\begin{tabular}[c]{@{}l@{}}Scene 1 \\ (ALS, GeoTIFF)\end{tabular}}                 & \begin{tabular}[c]{@{}l@{}}\SI{4712,1\pm613,8}{\second}\\ \SI{6570}{\mega\byte}\end{tabular} & \begin{tabular}[c]{@{}l@{}}\SI{2773,0\pm53,8}{\second}\\ \SI{2246}{\mega\byte}\end{tabular} & \begin{tabular}[c]{@{}l@{}}\SI{2403,8\pm18,0}{\second}\\ \SI{2246}{\mega\byte}\end{tabular} & \begin{tabular}[c]{@{}l@{}}\SI{1912,1\pm20,22}{\second}\\ \SI{2246}{\mega\byte}\end{tabular} \\
\textbf{\begin{tabular}[c]{@{}l@{}}Scene 2 \\ (MLS, geometric\\  primitives)\end{tabular}} & \begin{tabular}[c]{@{}l@{}}\SI{68,0\pm2,3}{\second}\\ \SI{560}{\mega\byte}\end{tabular}      & \begin{tabular}[c]{@{}l@{}}\SI{23,9\pm0,2}{\second}\\ \SI{24}{\mega\byte}\end{tabular}      & \begin{tabular}[c]{@{}l@{}}\SI{23,0\pm0,3}{\second}\\ \SI{24}{\mega\byte}\end{tabular}      & \begin{tabular}[c]{@{}l@{}}\SI{16,1\pm0,6}{\second}\\ \SI{24}{\mega\byte}\end{tabular}       \\
\textbf{\begin{tabular}[c]{@{}l@{}}Scene 3 \\ (TLS, tree models)\end{tabular}}             & \begin{tabular}[c]{@{}l@{}}\SI{362,6\pm6,7}{\second}\\ \SI{5360}{\mega\byte}\end{tabular}    & \begin{tabular}[c]{@{}l@{}}\SI{97,2\pm1,3}{\second}\\ \SI{314}{\mega\byte}\end{tabular}     & \begin{tabular}[c]{@{}l@{}}\SI{74,4\pm1,0}{\second}\\ \SI{314}{\mega\byte}\end{tabular}     & \begin{tabular}[c]{@{}l@{}}\SI{62,4\pm0,3}{\second}\\ \SI{314}{\mega\byte}\end{tabular}     
\end{tabular}

\cprotect\caption{Performance comparison of HELIOS and HELIOS++ with different options. We use default parameters for waveform modelling (\verb|beamSampleQuality=3| and \verb|binSize_ns=0.25|; \verb|numBins=100| and \verb|numFullwaveBins=200| for HELIOS++ and HELIOS, respectively). Runtimes are average of three runs ($\pm$ standard deviation), memory footprint is the highest value (maximum) during the full run.}
\label{tab:performance}
\end{sidewaystable}

 In the case with largest improvement over HELIOS, runtimes of HELIOS++ are lower by 83~\% (TLS) while using only 6~\% of the previously required memory. Especially for large scenes, the issue of memory footprint has been a limiting factor for the usability of HELIOS. In the case of the ALS scene, the memory consumption can be reduced by 66~\%, from more than \SI{6}{\giga\byte} to just above \SI{2}{\giga\byte}, with a runtime reduction of 51~\%. For the MLS scene using the simple geometric shapes, the reduced processing overhead leads to a reduction of 96~\% in memory usage and 76~\% in runtime. From these results we deduce a significant improvement both in runtime and memory footprint when comparing any configuration of HELIOS++ to the previous Java version.

\section{Conclusions}
\label{sec:conclusions}
With HELIOS++, we present an open-source laser scanning simulation framework that enables highly performant virtual laser scanning (VLS). In its C++ implementation and with the possibility to use the \verb|pyhelios| package in Python to manipulate simulation parameters, we opt for an efficient and easy-to-use software. While physical accuracy and realism may be superseded by complementing simulation software, HELIOS++ provides a flexible solution to balance computational requirements (runtime, memory footprint) and quality of results (physical realism), while being easy to use and configure for users.

Different studies using the HELIOS concept demonstrate the usefulness of virtually acquired laser scanning data for analyses in different categories of purpose. The main categories are (a) planning of flight patterns (ULS, ALS) or scan positions (TLS); (b) generation of ground-truth data for validation of algorithms that extract parameters (i.a. for forestry)  from point clouds, (c) generation of training data for supervised machine learning, and (d) testing and development of novel sensors and algorithms.

The novel object model of HELIOS++, the transmissive voxel, allows a stochastic simulation of penetrable objects, such as vegetation canopy, without the need for a highly detailed 3D mesh model. Generally, HELIOS++ performs up to of $5.8\,\times$ faster than its predecessor HELIOS on common scenes, and allows the user to omit intensive calculations if not required (e.g. the calculation of the echo width for returned pulses). 

To summarise, HELIOS++ is a versatile, easy-to-use, well documented scientific software for virtual laser scanning simulations that provides a tool to generate VLS data, thereby complementing data obtained by real-world laser scanning. The framework invites users to experiment and develop new ideas, while being able to rely on established algorithms from the literature.

\subsubsection*{Acknowledgements}
The authors wish to acknowledge the contribution of Patrick Herbers (Ruhr-Universität Bochum), for the improvement the triangle-ray intersection in C++, which helped to significantly reduce the runtime of HELIOS++.

Figures~\ref{fig:geile_figure}, \ref{fig:geile_results}, \ref{fig:scan_patterns}, and \ref{fig:als_demo} show an airplane model CC-BY Emmanuel Beranger. Figures ~\ref{fig:geile_figure} and \ref{fig:geile_results} show a house model by \url{free3d.com} user \verb|gerhald3d|, and a drone model by \url{cgtrader.com} user \verb|CGaxr|. 

\bibliography{mybibfile}

\begin{thebibliography}{57}
\expandafter\ifx\csname natexlab\endcsname\relax\def\natexlab#1{#1}\fi
\providecommand{\url}[1]{\texttt{#1}}
\providecommand{\href}[2]{#2}
\providecommand{\path}[1]{#1}
\providecommand{\DOIprefix}{doi:}
\providecommand{\ArXivprefix}{arXiv:}
\providecommand{\URLprefix}{URL: }
\providecommand{\Pubmedprefix}{pmid:}
\providecommand{\doi}[1]{\href{http://dx.doi.org/#1}{\path{#1}}}
\providecommand{\Pubmed}[1]{\href{pmid:#1}{\path{#1}}}
\providecommand{\bibinfo}[2]{#2}
\ifx\xfnm\relax \def\xfnm[#1]{\unskip,\space#1}\fi
%Type = Misc
\bibitem[{{ASPRS}(2011)}]{asprs2011}
\bibinfo{author}{{ASPRS}}, \bibinfo{year}{2011}.
\newblock \bibinfo{title}{{LAS} specification}.
\newblock
  \bibinfo{note}{\url{https://www.asprs.org/wp-content/uploads/2010/12/LAS\_1\_4\_r13.pdf}}.
%Type = Article
\bibitem[{Backes et~al.(2020)Backes, Smigaj, Schimka, Zahs, Grznárová and
  Scaioni}]{Backes_2020}
\bibinfo{author}{Backes, D.}, \bibinfo{author}{Smigaj, M.},
  \bibinfo{author}{Schimka, M.}, \bibinfo{author}{Zahs, V.},
  \bibinfo{author}{Grznárová, A.}, \bibinfo{author}{Scaioni, M.},
  \bibinfo{year}{2020}.
\newblock \bibinfo{title}{{River Morphology Monitoring of a Small-Scale Alpine
  Riverbed using Drone Photogrammetry and LiDAR}}.
\newblock \bibinfo{journal}{ISPRS - International Archives of the
  Photogrammetry, Remote Sensing and Spatial Information Sciences}
  \bibinfo{volume}{XLIII-B2-2020}, \bibinfo{pages}{1017–1024}.
\newblock \DOIprefix\doi{10.5194/isprs-archives-XLIII-B2-2020-1017-2020}.
%Type = Inproceedings
\bibitem[{Bechtold et~al.(2016)Bechtold, Hämmerle and Höfle}]{Bechtold_2016b}
\bibinfo{author}{Bechtold, S.}, \bibinfo{author}{Hämmerle, M.},
  \bibinfo{author}{Höfle, B.}, \bibinfo{year}{2016}.
\newblock \bibinfo{title}{{Simulated full-waveform laser scanning of outcrops
  for development of point cloud analysis algorithms and survey planning: An
  application for the HELIOS LiDAR simulation framework}}, in:
  \bibinfo{booktitle}{{Proceedings of the 2nd Virtual Geoscience Conference,
  Bergen, Norway}}, pp. \bibinfo{pages}{57--58}.
\newblock \URLprefix
  \url{http://lvisa.geog.uni-heidelberg.de/papers/2016/Bechtold_et_al_2016.pdf}.
%Type = Article
\bibitem[{Bechtold and Höfle(2016)}]{Bechtold_2016}
\bibinfo{author}{Bechtold, S.}, \bibinfo{author}{Höfle, B.},
  \bibinfo{year}{2016}.
\newblock \bibinfo{title}{{HELIOS: A Multi-Purpose LiDAR Simulation Framework
  for Research, Planning and Training of Laser Scanning Operations with
  Airborne, Ground-Based Mobile and Stationary Platforms}}.
\newblock \bibinfo{journal}{ISPRS Annals of Photogrammetry, Remote Sensing and
  Spatial Information Sciences} \bibinfo{volume}{III–3},
  \bibinfo{pages}{161–168}.
\newblock \DOIprefix\doi{10.5194/isprs-annals-III-3-161-2016}.
%Type = Article
\bibitem[{Bentley(1975)}]{Bentley.1975}
\bibinfo{author}{Bentley, J.L.}, \bibinfo{year}{1975}.
\newblock \bibinfo{title}{{Multidimensional binary search trees used for
  associative searching}}.
\newblock \bibinfo{journal}{Communications of the ACM} \bibinfo{volume}{18},
  \bibinfo{pages}{509--517}.
\newblock \DOIprefix\doi{https://doi.org/10.1145/361002.361007}.
%Type = Article
\bibitem[{Calders et~al.(2013)Calders, Lewis, Disney, Verbesselt and
  Herold}]{Calders.2013}
\bibinfo{author}{Calders, K.}, \bibinfo{author}{Lewis, P.},
  \bibinfo{author}{Disney, M.}, \bibinfo{author}{Verbesselt, J.},
  \bibinfo{author}{Herold, M.}, \bibinfo{year}{2013}.
\newblock \bibinfo{title}{{Investigating assumptions of crown archetypes for
  modelling LiDAR returns}}.
\newblock \bibinfo{journal}{{Remote Sensing of Environment}}
  \bibinfo{volume}{134}, \bibinfo{pages}{39--49}.
\newblock \DOIprefix\doi{10.1016/j.rse.2013.02.018}.
%Type = Misc
\bibitem[{Carlsson et~al.(2001)Carlsson, Steinvall and
  Letalick}]{Carlsson.2001}
\bibinfo{author}{Carlsson, T.}, \bibinfo{author}{Steinvall, O.},
  \bibinfo{author}{Letalick, D.}, \bibinfo{year}{2001}.
\newblock \bibinfo{title}{{Signature simulation and signal analysis for 3-D
  laser radar}}.
%Type = Inproceedings
\bibitem[{{Danielczuk} et~al.(2019){Danielczuk}, {Matl}, {Gupta}, {Li}, {Lee},
  {Mahler} and {Goldberg}}]{Danielczuk2019}
\bibinfo{author}{{Danielczuk}, M.}, \bibinfo{author}{{Matl}, M.},
  \bibinfo{author}{{Gupta}, S.}, \bibinfo{author}{{Li}, A.},
  \bibinfo{author}{{Lee}, A.}, \bibinfo{author}{{Mahler}, J.},
  \bibinfo{author}{{Goldberg}, K.}, \bibinfo{year}{2019}.
\newblock \bibinfo{title}{{Segmenting Unknown 3D Objects from Real Depth Images
  using Mask R-CNN Trained on Synthetic Data}}, in: \bibinfo{booktitle}{2019
  International Conference on Robotics and Automation (ICRA)}, pp.
  \bibinfo{pages}{7283--7290}.
\newblock \DOIprefix\doi{10.1109/ICRA.2019.8793744}.
%Type = Article
\bibitem[{Disney et~al.(2000)Disney, Lewis and North}]{Disney.2000}
\bibinfo{author}{Disney, M.}, \bibinfo{author}{Lewis, P.},
  \bibinfo{author}{North, P.}, \bibinfo{year}{2000}.
\newblock \bibinfo{title}{{Monte Carlo ray tracing in optical canopy
  reflectance modelling}}.
\newblock \bibinfo{journal}{Remote Sensing Reviews} \bibinfo{volume}{18},
  \bibinfo{pages}{163--196}.
\newblock \DOIprefix\doi{10.1080/02757250009532389}.
%Type = Article
\bibitem[{Disney et~al.(2010)Disney, Kalogirou, Lewis, Prieto-Blanco, Hancock
  and Pfeifer}]{Disney.2010}
\bibinfo{author}{Disney, M.I.}, \bibinfo{author}{Kalogirou, V.},
  \bibinfo{author}{Lewis, P.}, \bibinfo{author}{Prieto-Blanco, A.},
  \bibinfo{author}{Hancock, S.}, \bibinfo{author}{Pfeifer, M.},
  \bibinfo{year}{2010}.
\newblock \bibinfo{title}{{Simulating the impact of discrete-return lidar
  system and survey characteristics over young conifer and broadleaf forests}}.
\newblock \bibinfo{journal}{{Remote Sensing of Environment}}
  \bibinfo{volume}{114}, \bibinfo{pages}{1546--1560}.
\newblock \DOIprefix\doi{10.1016/j.rse.2010.02.009}.
%Type = Article
\bibitem[{{Disney} et~al.(2009){Disney}, {Lewis}, {Bouvet}, {Prieto-Blanco} and
  {Hancock}}]{Disney.2009}
\bibinfo{author}{{Disney}, M.I.}, \bibinfo{author}{{Lewis}, P.E.},
  \bibinfo{author}{{Bouvet}, M.}, \bibinfo{author}{{Prieto-Blanco}, A.},
  \bibinfo{author}{{Hancock}, S.}, \bibinfo{year}{2009}.
\newblock \bibinfo{title}{{Quantifying Surface Reflectivity for Spaceborne
  LiDAR via Two Independent Methods}}.
\newblock \bibinfo{journal}{IEEE Transactions on Geoscience and Remote Sensing}
  \bibinfo{volume}{47}, \bibinfo{pages}{3262--3271}.
\newblock \DOIprefix\doi{10.1109/TGRS.2009.2019268}.
%Type = Article
\bibitem[{Gastellu-Etchegorry et~al.(2015)Gastellu-Etchegorry, Yin, Lauret,
  Cajgfinger, Gregoire, Grau, Feret, Lopes, Guilleux, Dedieu, Malenovsk{\'y},
  Cook, Morton, Rubio, Durrieu, Cazanave, Martin and
  Ristorcelli}]{GastelluEtchegorry.2015}
\bibinfo{author}{Gastellu-Etchegorry, J.P.}, \bibinfo{author}{Yin, T.},
  \bibinfo{author}{Lauret, N.}, \bibinfo{author}{Cajgfinger, T.},
  \bibinfo{author}{Gregoire, T.}, \bibinfo{author}{Grau, E.},
  \bibinfo{author}{Feret, J.B.}, \bibinfo{author}{Lopes, M.},
  \bibinfo{author}{Guilleux, J.}, \bibinfo{author}{Dedieu, G.},
  \bibinfo{author}{Malenovsk{\'y}, Z.}, \bibinfo{author}{Cook, B.},
  \bibinfo{author}{Morton, D.}, \bibinfo{author}{Rubio, J.},
  \bibinfo{author}{Durrieu, S.}, \bibinfo{author}{Cazanave, G.},
  \bibinfo{author}{Martin, E.}, \bibinfo{author}{Ristorcelli, T.},
  \bibinfo{year}{2015}.
\newblock \bibinfo{title}{{Discrete Anisotropic Radiative Transfer (DART 5) for
  Modeling Airborne and Satellite Spectroradiometer and LIDAR Acquisitions of
  Natural and Urban Landscapes}}.
\newblock \bibinfo{journal}{{Remote Sensing}} \bibinfo{volume}{7},
  \bibinfo{pages}{1667--1701}.
\newblock \DOIprefix\doi{10.3390/rs70201667}.
%Type = Article
\bibitem[{Gastellu-Etchegorry et~al.(2016)Gastellu-Etchegorry, Yin, Lauret,
  Grau, Rubio, Cook, Morton and Sun}]{GastelluEtchegorry.2016}
\bibinfo{author}{Gastellu-Etchegorry, J.P.}, \bibinfo{author}{Yin, T.},
  \bibinfo{author}{Lauret, N.}, \bibinfo{author}{Grau, E.},
  \bibinfo{author}{Rubio, J.}, \bibinfo{author}{Cook, B.D.},
  \bibinfo{author}{Morton, D.C.}, \bibinfo{author}{Sun, G.},
  \bibinfo{year}{2016}.
\newblock \bibinfo{title}{{Simulation of satellite, airborne and terrestrial
  LiDAR with DART (I): Waveform simulation with quasi-Monte Carlo ray
  tracing}}.
\newblock \bibinfo{journal}{{Remote Sensing of Environment}}
  \bibinfo{volume}{184}, \bibinfo{pages}{418--435}.
\newblock \DOIprefix\doi{10.1016/j.rse.2016.07.010}.
%Type = Article
\bibitem[{Giannetti et~al.(2018)Giannetti, Puletti, Quatrini, Travaglini,
  Bottalico, Corona and Chirici}]{giannetti_2018}
\bibinfo{author}{Giannetti, F.}, \bibinfo{author}{Puletti, N.},
  \bibinfo{author}{Quatrini, V.}, \bibinfo{author}{Travaglini, D.},
  \bibinfo{author}{Bottalico, F.}, \bibinfo{author}{Corona, P.},
  \bibinfo{author}{Chirici, G.}, \bibinfo{year}{2018}.
\newblock \bibinfo{title}{{Integrating terrestrial and airborne laser scanning
  for the assessment of single-tree attributes in Mediterranean forest
  stands}}.
\newblock \bibinfo{journal}{European Journal of Remote Sensing}
  \bibinfo{volume}{51}, \bibinfo{pages}{795--807}.
\newblock \DOIprefix\doi{10.1080/22797254.2018.1482733}.
%Type = Article
\bibitem[{Golub and Kahan(1965)}]{Golub.1965}
\bibinfo{author}{Golub, G.}, \bibinfo{author}{Kahan, W.}, \bibinfo{year}{1965}.
\newblock \bibinfo{title}{{Calculating the Singular Values and Pseudo-Inverse
  of a Matrix}}.
\newblock \bibinfo{journal}{Journal of the Society for Industrial and Applied
  Mathematics Series B Numerical Analysis} \bibinfo{volume}{2},
  \bibinfo{pages}{205--224}.
\newblock \DOIprefix\doi{https://doi.org/10.1137/0702016}.
%Type = Article
\bibitem[{Goodwin et~al.(2007)Goodwin, Coops and Culvenor}]{Goodwin.2007}
\bibinfo{author}{Goodwin, N.R.}, \bibinfo{author}{Coops, N.C.},
  \bibinfo{author}{Culvenor, D.S.}, \bibinfo{year}{2007}.
\newblock \bibinfo{title}{{Development of a simulation model to predict LiDAR
  interception in forested environments}}.
\newblock \bibinfo{journal}{{Remote Sensing of Environment}}
  \bibinfo{volume}{111}, \bibinfo{pages}{481--492}.
\newblock \DOIprefix\doi{10.1016/j.rse.2007.04.001}.
%Type = Article
\bibitem[{Hodge(2010)}]{Hodge.2010}
\bibinfo{author}{Hodge, R.A.}, \bibinfo{year}{2010}.
\newblock \bibinfo{title}{{Using simulated Terrestrial Laser Scanning to
  analyse errors in high-resolution scan data of irregular surfaces}}.
\newblock \bibinfo{journal}{{ISPRS Journal of Photogrammetry and Remote
  Sensing}} \bibinfo{volume}{65}, \bibinfo{pages}{227--240}.
\newblock \DOIprefix\doi{10.1016/j.isprsjprs.2010.01.001}.
%Type = Article
\bibitem[{Holmgren et~al.(2003)Holmgren, Nilsson and Olsson}]{Holmgren.2003}
\bibinfo{author}{Holmgren, J.}, \bibinfo{author}{Nilsson, M.},
  \bibinfo{author}{Olsson, H.}, \bibinfo{year}{2003}.
\newblock \bibinfo{title}{{Simulating the effects of LiDAR scanning angle for
  estimation of mean tree height and canopy closure}}.
\newblock \bibinfo{journal}{{Canadian Journal of Remote Sensing}}
  \bibinfo{volume}{29}, \bibinfo{pages}{623--632}.
\newblock \DOIprefix\doi{10.5589/m03-030}.
%Type = Article
\bibitem[{Hämmerle et~al.(2017)Hämmerle, Lukač, Chen, Koma, Wang, Anders and
  Höfle}]{Hammerle_2017}
\bibinfo{author}{Hämmerle, M.}, \bibinfo{author}{Lukač, N.},
  \bibinfo{author}{Chen, K.C.}, \bibinfo{author}{Koma, Z.},
  \bibinfo{author}{Wang, C.K.}, \bibinfo{author}{Anders, K.},
  \bibinfo{author}{Höfle, B.}, \bibinfo{year}{2017}.
\newblock \bibinfo{title}{{Simulating Various Terrestrial and UAV LiDAR
  Scanning Configurations for Understory Forest Structure Modelling}}.
\newblock \bibinfo{journal}{ISPRS Annals of Photogrammetry, Remote Sensing and
  Spatial Information Sciences} \bibinfo{volume}{IV-2/W4},
  \bibinfo{pages}{59–65}.
\newblock \DOIprefix\doi{10.5194/isprs-annals-IV-2-W4-59-2017}.
%Type = Article
\bibitem[{Isenburg(2013)}]{isenburg2013laszip}
\bibinfo{author}{Isenburg, M.}, \bibinfo{year}{2013}.
\newblock \bibinfo{title}{{LASzip: lossless compression of LiDAR data}}.
\newblock \bibinfo{journal}{Photogrammetric Engineering and Remote Sensing}
  \bibinfo{volume}{79}, \bibinfo{pages}{209--217}.
\newblock \URLprefix
  \url{https://www.cs.unc.edu/~isenburg/lastools/download/laszip.pdf}.
%Type = Article
\bibitem[{Kim et~al.(2012)Kim, Lee and Lee}]{Kim.2012}
\bibinfo{author}{Kim, S.}, \bibinfo{author}{Lee, I.}, \bibinfo{author}{Lee,
  M.}, \bibinfo{year}{2012}.
\newblock \bibinfo{title}{{LiDAR Waveform Simulation over Complex Targets}}.
\newblock \bibinfo{journal}{{ISPRS - International Archives of the
  Photogrammetry, Remote Sensing and Spatial Information Sciences}}
  \bibinfo{volume}{XXXIX-B7}, \bibinfo{pages}{517--522}.
\newblock \DOIprefix\doi{10.5194/isprsarchives-XXXIX-B7-517-2012}.
%Type = Inproceedings
\bibitem[{Kim et~al.(2009)Kim, Min, Kim, Lee and Jun}]{Kim.2009}
\bibinfo{author}{Kim, S.}, \bibinfo{author}{Min, S.}, \bibinfo{author}{Kim,
  G.}, \bibinfo{author}{Lee, I.}, \bibinfo{author}{Jun, C.},
  \bibinfo{year}{2009}.
\newblock \bibinfo{title}{{Data simulation of an airborne lidar system}}, in:
  \bibinfo{editor}{Turner, M.D.}, \bibinfo{editor}{Kamerman, G.W.} (Eds.),
  \bibinfo{booktitle}{{Laser Radar Technology and Applications XIV}},
  \bibinfo{publisher}{SPIE}. p. \bibinfo{pages}{73230C}.
\newblock \DOIprefix\doi{10.1117/12.818545}.
%Type = Article
\bibitem[{Kukko and Hyypp{\"a}(2009)}]{Kukko.2009}
\bibinfo{author}{Kukko, A.}, \bibinfo{author}{Hyypp{\"a}, J.},
  \bibinfo{year}{2009}.
\newblock \bibinfo{title}{{Small-footprint Laser Scanning Simulator for System
  Validation, Error Assessment, and Algorithm Development}}.
\newblock \bibinfo{journal}{{Photogrammetric Engineering {\&} Remote Sensing}}
  \bibinfo{volume}{75}, \bibinfo{pages}{1177--1189}.
\newblock \DOIprefix\doi{10.14358/PERS.75.10.1177}.
%Type = Article
\bibitem[{Lewis(1999)}]{Lewis.1999}
\bibinfo{author}{Lewis, P.}, \bibinfo{year}{1999}.
\newblock \bibinfo{title}{{Three-dimensional plant modelling for remote sensing
  simulation studies using the Botanical Plant Modelling System}}.
\newblock \bibinfo{journal}{{Agronomie}} \bibinfo{volume}{19},
  \bibinfo{pages}{185--210}.
\newblock \DOIprefix\doi{10.1051/agro:19990302}.
%Type = Article
\bibitem[{Lewis and Muller(1993)}]{Lewis.1993}
\bibinfo{author}{Lewis, P.}, \bibinfo{author}{Muller, J.P.},
  \bibinfo{year}{1993}.
\newblock \bibinfo{title}{The advanced radiometric ray tracer: Ararat for plant
  canopy reflectance simulation}.
\newblock \bibinfo{journal}{Proc. 29th Conf. Int. Soc. Photogramm. Remote
  Sens.} \bibinfo{volume}{29}.
\newblock \URLprefix
  \url{http://www.isprs.org/proceedings/XXIX/congress/part7/26_XXIX-part7.pdf}.
%Type = Article
\bibitem[{Li et~al.(2020)Li, Mu, Soma, Wan, Qi, Hu, Zhang, Tong and
  Yan}]{Li_Mu_Soma_Wan_Qi_Hu_Zhang_Tong_Yan_2020}
\bibinfo{author}{Li, L.}, \bibinfo{author}{Mu, X.}, \bibinfo{author}{Soma, M.},
  \bibinfo{author}{Wan, P.}, \bibinfo{author}{Qi, J.}, \bibinfo{author}{Hu,
  R.}, \bibinfo{author}{Zhang, W.}, \bibinfo{author}{Tong, Y.},
  \bibinfo{author}{Yan, G.}, \bibinfo{year}{2020}.
\newblock \bibinfo{title}{{An Iterative-Mode Scan Design of Terrestrial Laser
  Scanning in Forests for Minimizing Occlusion Effects}}.
\newblock \bibinfo{journal}{IEEE Transactions on Geoscience and Remote Sensing}
  , \bibinfo{pages}{1–20}\DOIprefix\doi{10.1109/TGRS.2020.3018643}.
%Type = Article
\bibitem[{Pires~de Lima and Marfurt(2019)}]{Pires_de_Lima_2019}
\bibinfo{author}{Pires~de Lima, R.}, \bibinfo{author}{Marfurt, K.},
  \bibinfo{year}{2019}.
\newblock \bibinfo{title}{{Convolutional Neural Network for Remote-Sensing
  Scene Classification: Transfer Learning Analysis}}.
\newblock \bibinfo{journal}{Remote Sensing} \bibinfo{volume}{12},
  \bibinfo{pages}{86}.
\newblock \URLprefix \url{http://dx.doi.org/10.3390/rs12010086},
  \DOIprefix\doi{10.3390/rs12010086}.
%Type = Inproceedings
\bibitem[{Lin and Wang(2019)}]{Lin_2019}
\bibinfo{author}{Lin, C.H.}, \bibinfo{author}{Wang, C.K.},
  \bibinfo{year}{2019}.
\newblock \bibinfo{title}{{Point Density Simulation for ALS Survey}}, in:
  \bibinfo{booktitle}{Proceedings of the 11th International Conference on
  Mobile Mapping Technology (MMT2019)}, pp. \bibinfo{pages}{157--160}.
\newblock \URLprefix
  \url{https://www.geog.uni-heidelberg.de/md/chemgeo/geog/gis/mmt2019-lin_and_wang_compr.pdf}.
%Type = Article
\bibitem[{Liu et~al.(2019a)Liu, Skidmore, Wang, Zhu, Premier, Heurich, Beudert
  and Jones}]{Liu_Skidmore_Wang_Zhu_Premier_Heurich_Beudert_Jones_2019}
\bibinfo{author}{Liu, J.}, \bibinfo{author}{Skidmore, A.K.},
  \bibinfo{author}{Wang, T.}, \bibinfo{author}{Zhu, X.},
  \bibinfo{author}{Premier, J.}, \bibinfo{author}{Heurich, M.},
  \bibinfo{author}{Beudert, B.}, \bibinfo{author}{Jones, S.},
  \bibinfo{year}{2019}a.
\newblock \bibinfo{title}{{Variation of leaf angle distribution quantified by
  terrestrial LiDAR in natural European beech forest}}.
\newblock \bibinfo{journal}{ISPRS Journal of Photogrammetry and Remote Sensing}
  \bibinfo{volume}{148}, \bibinfo{pages}{208–220}.
\newblock \DOIprefix\doi{10.1016/j.isprsjprs.2019.01.005}.
%Type = Article
\bibitem[{Liu et~al.(2019b)Liu, Wang, Skidmore, Jones, Heurich, Beudert and
  Premier}]{Liu_Wang_Skidmore_Jones_Heurich_Beudert_Premier_2019}
\bibinfo{author}{Liu, J.}, \bibinfo{author}{Wang, T.},
  \bibinfo{author}{Skidmore, A.K.}, \bibinfo{author}{Jones, S.},
  \bibinfo{author}{Heurich, M.}, \bibinfo{author}{Beudert, B.},
  \bibinfo{author}{Premier, J.}, \bibinfo{year}{2019}b.
\newblock \bibinfo{title}{{Comparison of terrestrial LiDAR and digital
  hemispherical photography for estimating leaf angle distribution in European
  broadleaf beech forests}}.
\newblock \bibinfo{journal}{ISPRS Journal of Photogrammetry and Remote Sensing}
  \bibinfo{volume}{158}, \bibinfo{pages}{76–89}.
\newblock \DOIprefix\doi{10.1016/j.isprsjprs.2019.09.015}.
%Type = Inproceedings
\bibitem[{Lohani and Mishra(2007)}]{Lohani.2007}
\bibinfo{author}{Lohani, B.}, \bibinfo{author}{Mishra, R.K.},
  \bibinfo{year}{2007}.
\newblock \bibinfo{title}{{Generating~LiDAR~Data in Laboratory:
  LiDAR~Simulator}}, in: \bibinfo{booktitle}{{ISPRS Workshop on Laser Scanning
  2007 and SilviLaser 2007}}, pp. \bibinfo{pages}{264--269}.
\newblock \URLprefix
  \url{https://www.isprs.org/proceedings/XXXVI/3-W52/final_papers/Lohani_2007.pdf}.
%Type = Article
\bibitem[{Lovell et~al.(2005)Lovell, Jupp, Newnham, Coops and
  Culvenor}]{Lovell.2005}
\bibinfo{author}{Lovell, J.L.}, \bibinfo{author}{Jupp, D.},
  \bibinfo{author}{Newnham, G.J.}, \bibinfo{author}{Coops, N.C.},
  \bibinfo{author}{Culvenor, D.S.}, \bibinfo{year}{2005}.
\newblock \bibinfo{title}{{Simulation study for finding optimal LiDAR
  acquisition parameters for forest height retrieval}}.
\newblock \bibinfo{journal}{{Forest Ecology and Management}}
  \bibinfo{volume}{214}, \bibinfo{pages}{398--412}.
\newblock \DOIprefix\doi{10.1016/j.foreco.2004.07.077}.
%Type = Article
\bibitem[{Martínez~Sánchez et~al.(2019)Martínez~Sánchez, Váquez~Alvarez,
  López~Vilarino, Fernández~Rivera, Cabaleiro~Domínguez and
  Fernández~Pena}]{Martinez_2019}
\bibinfo{author}{Martínez~Sánchez, J.}, \bibinfo{author}{Váquez~Alvarez,
  A.}, \bibinfo{author}{López~Vilarino, D.},
  \bibinfo{author}{Fernández~Rivera, F.},
  \bibinfo{author}{Cabaleiro~Domínguez, J.C.},
  \bibinfo{author}{Fernández~Pena, T.}, \bibinfo{year}{2019}.
\newblock \bibinfo{title}{{Fast Ground Filtering of Airborne LiDAR Data Based
  on Iterative Scan-Line Spline Interpolation}}.
\newblock \bibinfo{journal}{Remote Sensing} \bibinfo{volume}{11},
  \bibinfo{pages}{2256}.
\newblock \DOIprefix\doi{10.3390/rs11192256}.
%Type = Misc
\bibitem[{Morsdorf et~al.(2007)Morsdorf, Frey, Koetz and Meier}]{Morsdorf.2007}
\bibinfo{author}{Morsdorf, F.}, \bibinfo{author}{Frey, O.},
  \bibinfo{author}{Koetz, B.}, \bibinfo{author}{Meier, E.},
  \bibinfo{year}{2007}.
\newblock \bibinfo{title}{{Ray Tracing for Modeling of Small Footprint Airborne
  Laser Scanning Returns}}.
\newblock \DOIprefix\doi{10.3929/ETHZ-B-000107380}.
%Type = Article
\bibitem[{North(1996)}]{North.1996}
\bibinfo{author}{North, P.}, \bibinfo{year}{1996}.
\newblock \bibinfo{title}{{Three-dimensional forest light interaction model
  using a Monte Carlo method}}.
\newblock \bibinfo{journal}{{IEEE Transactions on Geoscience and Remote
  Sensing}} \bibinfo{volume}{34}, \bibinfo{pages}{946--956}.
\newblock \DOIprefix\doi{10.1109/36.508411}.
%Type = Article
\bibitem[{North et~al.(2010)North, Rosette, Su{\'a}rez and Los}]{North.2010}
\bibinfo{author}{North, P.R.J.}, \bibinfo{author}{Rosette, J.A.B.},
  \bibinfo{author}{Su{\'a}rez, J.C.}, \bibinfo{author}{Los, S.O.},
  \bibinfo{year}{2010}.
\newblock \bibinfo{title}{{A Monte Carlo radiative transfer model of satellite
  waveform LiDAR}}.
\newblock \bibinfo{journal}{{International Journal of Remote Sensing}}
  \bibinfo{volume}{31}, \bibinfo{pages}{1343--1358}.
\newblock \DOIprefix\doi{10.1080/01431160903380664}.
%Type = Article
\bibitem[{Park et~al.(2020)Park, Baek, Dinare, Lee, Park, Ahn, Kim, Medina,
  Choi, Kim and et~al.}]{Park_2020}
\bibinfo{author}{Park, M.}, \bibinfo{author}{Baek, Y.},
  \bibinfo{author}{Dinare, M.}, \bibinfo{author}{Lee, D.},
  \bibinfo{author}{Park, K.H.}, \bibinfo{author}{Ahn, J.},
  \bibinfo{author}{Kim, D.}, \bibinfo{author}{Medina, J.},
  \bibinfo{author}{Choi, W.J.}, \bibinfo{author}{Kim, S.},
  \bibinfo{author}{et~al.}, \bibinfo{year}{2020}.
\newblock \bibinfo{title}{{Hetero-integration enables fast switching
  time-of-flight sensors for light detection and ranging}}.
\newblock \bibinfo{journal}{Scientific Reports} \bibinfo{volume}{10},
  \bibinfo{pages}{2764}.
\newblock \DOIprefix\doi{10.1038/s41598-020-59677-x}.
%Type = Inproceedings
\bibitem[{Previtali et~al.(2019)Previtali, Díaz-Vilariño, Scaioni and
  Frías~Nores}]{Previtali_2019}
\bibinfo{author}{Previtali, M.}, \bibinfo{author}{Díaz-Vilariño, L.},
  \bibinfo{author}{Scaioni, M.}, \bibinfo{author}{Frías~Nores, E.},
  \bibinfo{year}{2019}.
\newblock \bibinfo{title}{{Evaluation of the Expected Data Quality in Laser
  Scanning Surveying of Archaeological Sites}}, in: \bibinfo{booktitle}{{4th
  International Conference on Metrology for Archaeology and Cultural Heritage,
  Florence, Italy}}, pp. \bibinfo{pages}{19--24}.
\newblock \URLprefix
  \url{https://re.public.polimi.it/retrieve/handle/11311/1124569/476842/Previtali\%20et\%20al\%202019\%20MetroArchaeo.pdf}.
%Type = Inproceedings
\bibitem[{Qi et~al.(2017)Qi, Yi, Su and Guibas}]{qi2017pointnet++}
\bibinfo{author}{Qi, C.R.}, \bibinfo{author}{Yi, L.}, \bibinfo{author}{Su, H.},
  \bibinfo{author}{Guibas, L.J.}, \bibinfo{year}{2017}.
\newblock \bibinfo{title}{{PointNet++}: Deep hierarchical feature learning on
  point sets in a metric space}, in: \bibinfo{booktitle}{Advances in Neural
  Information Processing Systems}, pp. \bibinfo{pages}{5099--5108}.
%Type = Misc
\bibitem[{Ramey(2004)}]{boostserial}
\bibinfo{author}{Ramey, R.}, \bibinfo{year}{2004}.
\newblock \bibinfo{title}{{C++ BOOST Serialization}}.
\newblock \URLprefix
  \url{https://www.boost.org/doc/libs/1_72_0/libs/serialization/doc/index.html}.
%Type = Article
\bibitem[{Ranson and Sun(2000)}]{Ranson.2000}
\bibinfo{author}{Ranson, K.J.}, \bibinfo{author}{Sun, G.},
  \bibinfo{year}{2000}.
\newblock \bibinfo{title}{{Modeling LiDAR Returns from Forest Canopies}}.
\newblock \bibinfo{journal}{{IEEE Transactions on Geoscience and Remote
  Sensing}} \bibinfo{volume}{38}, \bibinfo{pages}{2617--2626}.
\newblock \DOIprefix\doi{10.1109/36.885208}.
%Type = Article
\bibitem[{Rebolj et~al.(2017)Rebolj, Pučko, Babič, Bizjak and
  Mongus}]{Rebolj2017}
\bibinfo{author}{Rebolj, D.}, \bibinfo{author}{Pučko, Z.},
  \bibinfo{author}{Babič, N.C.}, \bibinfo{author}{Bizjak, M.},
  \bibinfo{author}{Mongus, D.}, \bibinfo{year}{2017}.
\newblock \bibinfo{title}{{Point cloud quality requirements for Scan-vs-BIM
  based automated construction progress monitoring}}.
\newblock \bibinfo{journal}{Automation in Construction} \bibinfo{volume}{84},
  \bibinfo{pages}{323–334}.
\newblock \DOIprefix\doi{10.1016/j.autcon.2017.09.021}.
%Type = Article
\bibitem[{Schlager et~al.(2020)Schlager, Muckenhuber, Schmidt, Holzer, Rott,
  Maier, Saad, Kirchengast, Stettinger, Watzenig and Ruebsam}]{Schlager_2020}
\bibinfo{author}{Schlager, B.}, \bibinfo{author}{Muckenhuber, S.},
  \bibinfo{author}{Schmidt, S.}, \bibinfo{author}{Holzer, H.},
  \bibinfo{author}{Rott, R.}, \bibinfo{author}{Maier, F.M.},
  \bibinfo{author}{Saad, K.}, \bibinfo{author}{Kirchengast, M.},
  \bibinfo{author}{Stettinger, G.}, \bibinfo{author}{Watzenig, D.},
  \bibinfo{author}{Ruebsam, J.}, \bibinfo{year}{2020}.
\newblock \bibinfo{title}{{State-of-the-Art Sensor Models for Virtual Testing
  of Advanced Driver Assistance Systems/Autonomous Driving Functions}}.
\newblock \bibinfo{journal}{SAE International Journal of Connected and
  Automated Vehicles} \bibinfo{volume}{3}, \bibinfo{pages}{233–261}.
\newblock \DOIprefix\doi{10.4271/12-03-03-0018}.
%Type = Inproceedings
\bibitem[{Schäfer et~al.(2019)Schäfer, Faßnacht, Höfle and
  Weiser}]{Schaefer_2020}
\bibinfo{author}{Schäfer, J.}, \bibinfo{author}{Faßnacht, F.},
  \bibinfo{author}{Höfle, B.}, \bibinfo{author}{Weiser, H.},
  \bibinfo{year}{2019}.
\newblock \bibinfo{title}{{Das SYSSIFOSS-Projekt: Synthetische
  3D-Fernerkundungsdaten für verbesserte Waldinventurmodelle}}, in:
  \bibinfo{booktitle}{{2. Symposium zur angewandten Satellitenerdbeoachtung,
  Cologne, Germany}}.
\newblock \URLprefix
  \url{https://www.dialogplattform-erdbeobachtung.de/downloads/cms/Stele3/EO-Symposium_Jannika.Schaefer.pdf}.
%Type = Article
\bibitem[{Tulldahl and Steinvall(1999)}]{Tulldahl.1999}
\bibinfo{author}{Tulldahl, H.M.}, \bibinfo{author}{Steinvall, K.O.},
  \bibinfo{year}{1999}.
\newblock \bibinfo{title}{{Analytical waveform generation from small objects in
  LiDAR bathymetry}}.
\newblock \bibinfo{journal}{{Applied optics}} \bibinfo{volume}{38},
  \bibinfo{pages}{1021--1039}.
\newblock \DOIprefix\doi{10.1364/ao.38.001021}.
%Type = Article
\bibitem[{Vincent et~al.(2017)Vincent, Antin, Laurans, Heurtebize, Durrieu,
  Lavalley and Dauzat}]{vincent_2017}
\bibinfo{author}{Vincent, G.}, \bibinfo{author}{Antin, C.},
  \bibinfo{author}{Laurans, M.}, \bibinfo{author}{Heurtebize, J.},
  \bibinfo{author}{Durrieu, S.}, \bibinfo{author}{Lavalley, C.},
  \bibinfo{author}{Dauzat, J.}, \bibinfo{year}{2017}.
\newblock \bibinfo{title}{{Mapping plant area index of tropical evergreen
  forest by airborne laser scanning. A cross-validation study using LAI2200
  optical sensor}}.
\newblock \bibinfo{journal}{Remote Sensing of Environment}
  \bibinfo{volume}{198}, \bibinfo{pages}{254 -- 266}.
\newblock \URLprefix
  \url{http://www.sciencedirect.com/science/article/pii/S003442571730233X},
  \DOIprefix\doi{https://doi.org/10.1016/j.rse.2017.05.034}.
%Type = Article
\bibitem[{Wang(2020)}]{Wang_2020}
\bibinfo{author}{Wang, D.}, \bibinfo{year}{2020}.
\newblock \bibinfo{title}{{Unsupervised semantic and instance segmentation of
  forest point clouds}}.
\newblock \bibinfo{journal}{ISPRS Journal of Photogrammetry and Remote Sensing}
  \bibinfo{volume}{165}, \bibinfo{pages}{86–97}.
\newblock \DOIprefix\doi{10.1016/j.isprsjprs.2020.04.020}.
%Type = Article
\bibitem[{Wang et~al.(2020)Wang, Schraik, Hovi and
  Rautiainen}]{Wang_Schraik_Hovi_Rautiainen_2020}
\bibinfo{author}{Wang, D.}, \bibinfo{author}{Schraik, D.},
  \bibinfo{author}{Hovi, A.}, \bibinfo{author}{Rautiainen, M.},
  \bibinfo{year}{2020}.
\newblock \bibinfo{title}{{Direct estimation of photon recollision probability
  using terrestrial laser scanning}}.
\newblock \bibinfo{journal}{Remote Sensing of Environment}
  \bibinfo{volume}{247}, \bibinfo{pages}{111932}.
\newblock \DOIprefix\doi{10.1016/j.rse.2020.111932}.
%Type = Inproceedings
\bibitem[{Wang et~al.(2013)Wang, Xie, Yan, Zhang and Mu}]{Wang.July2013}
\bibinfo{author}{Wang, Y.}, \bibinfo{author}{Xie, D.}, \bibinfo{author}{Yan,
  G.}, \bibinfo{author}{Zhang, W.}, \bibinfo{author}{Mu, X.},
  \bibinfo{year}{2013}.
\newblock \bibinfo{title}{{Analysis on the inversion accuracy of LAI based on
  simulated point clouds of terrestrial LiDAR of tree by ray tracing
  algorithm}}, in: \bibinfo{booktitle}{{IGARSS 2013-2013 IEEE International
  Geoscience and Remote Sensing Symposium}}, \bibinfo{publisher}{IEEE},
  \bibinfo{address}{Piscataway}. pp. \bibinfo{pages}{532--535}.
\newblock \DOIprefix\doi{10.1109/IGARSS.2013.6721210}.
%Type = Manual
\bibitem[{{Wavefront Technologies}(1992)}]{obj}
\bibinfo{author}{{Wavefront Technologies}}, \bibinfo{year}{1992}.
\newblock \bibinfo{title}{{Appendix B1. Object Files (.obj), Advanced
  Visualizer Manual}}.
\newblock \URLprefix
  \url{http://fegemo.github.io/cefet-cg/attachments/obj-spec.pdf}.
%Type = Inproceedings
\bibitem[{Weber and Penn(1995)}]{Weber.1995}
\bibinfo{author}{Weber, J.}, \bibinfo{author}{Penn, J.}, \bibinfo{year}{1995}.
\newblock \bibinfo{title}{{Creation and rendering of realistic trees}}, in:
  \bibinfo{editor}{Mair, S.G.}, \bibinfo{editor}{Cook, R.} (Eds.),
  \bibinfo{booktitle}{{Computer graphics proceedings}},
  \bibinfo{publisher}{{Association for Computing Machinery}},
  \bibinfo{address}{New York, NY}. pp. \bibinfo{pages}{119--128}.
\newblock \DOIprefix\doi{10.1145/218380.218427}.
%Type = Unpublished
\bibitem[{Weiser et~al.(2021)Weiser, Winiwarter, Fassnacht and
  Höfle}]{Weiser2020}
\bibinfo{author}{Weiser, H.}, \bibinfo{author}{Winiwarter, L.},
  \bibinfo{author}{Fassnacht, F.}, \bibinfo{author}{Höfle, B.},
  \bibinfo{year}{2021}.
\newblock \bibinfo{title}{{Opaque Voxel-based Tree Models for Virtual Laser
  Scanning in Forestry Applications}}.
\newblock \bibinfo{note}{In preparation}.
%Type = Article
\bibitem[{Winiwarter et~al.(2019)Winiwarter, Mandlburger, Schmohl and
  Pfeifer}]{Winiwarter2019}
\bibinfo{author}{Winiwarter, L.}, \bibinfo{author}{Mandlburger, G.},
  \bibinfo{author}{Schmohl, S.}, \bibinfo{author}{Pfeifer, N.},
  \bibinfo{year}{2019}.
\newblock \bibinfo{title}{{Classification of ALS Point Clouds Using End-to-End
  Deep Learning}}.
\newblock \bibinfo{journal}{PFG – Journal of Photogrammetry, Remote Sensing
  and Geoinformation Science} \URLprefix
  \url{https://doi.org/10.1007/s41064-019-00073-0},
  \DOIprefix\doi{10.1007/s41064-019-00073-0}.
%Type = Misc
\bibitem[{Winiwarter et~al.(2021)Winiwarter, Pena, Weiser, Anders, Saches,
  Searle and Höfle}]{lukas_winiwarter_2021_4452871}
\bibinfo{author}{Winiwarter, L.}, \bibinfo{author}{Pena, A.M.E.},
  \bibinfo{author}{Weiser, H.}, \bibinfo{author}{Anders, K.},
  \bibinfo{author}{Saches, J.M.}, \bibinfo{author}{Searle, M.},
  \bibinfo{author}{Höfle, B.}, \bibinfo{year}{2021}.
\newblock \bibinfo{title}{3dgeo-heidelberg/helios: Version 1.0.0}.
\newblock \URLprefix \url{https://doi.org/10.5281/zenodo.4452871},
  \DOIprefix\doi{10.5281/zenodo.4452871}.
%Type = Article
\bibitem[{Xiao et~al.(2019)Xiao, Zaforemska, Smigaj, Wang and
  Gaulton}]{Xiao_Zaforemska_Smigaj_Wang_Gaulton_2019}
\bibinfo{author}{Xiao, W.}, \bibinfo{author}{Zaforemska, A.},
  \bibinfo{author}{Smigaj, M.}, \bibinfo{author}{Wang, Y.},
  \bibinfo{author}{Gaulton, R.}, \bibinfo{year}{2019}.
\newblock \bibinfo{title}{Mean shift segmentation assessment for individual
  forest tree delineation from airborne lidar data}.
\newblock \bibinfo{journal}{Remote Sensing} \bibinfo{volume}{11},
  \bibinfo{pages}{1263}.
\newblock \DOIprefix\doi{10.3390/rs11111263}.
%Type = Article
\bibitem[{Zhang et~al.(2019)Zhang, Li, Guo, Yang and
  Wang}]{Zhang_Li_Guo_Yang_Wang_2019}
\bibinfo{author}{Zhang, Z.}, \bibinfo{author}{Li, J.}, \bibinfo{author}{Guo,
  Y.}, \bibinfo{author}{Yang, C.}, \bibinfo{author}{Wang, C.},
  \bibinfo{year}{2019}.
\newblock \bibinfo{title}{{3D Highway Curve Reconstruction From Mobile Laser
  Scanning Point Clouds}}.
\newblock \bibinfo{journal}{IEEE Transactions on Intelligent Transportation
  Systems} , \bibinfo{pages}{1–11}\DOIprefix\doi{10.1109/TITS.2019.2946259}.
%Type = Article
\bibitem[{Zhu et~al.(2020)Zhu, Liu, Skidmore, Premier and
  Heurich}]{Zhu_Liu_Skidmore_Premier_Heurich_2020}
\bibinfo{author}{Zhu, X.}, \bibinfo{author}{Liu, J.},
  \bibinfo{author}{Skidmore, A.K.}, \bibinfo{author}{Premier, J.},
  \bibinfo{author}{Heurich, M.}, \bibinfo{year}{2020}.
\newblock \bibinfo{title}{{A voxel matching method for effective leaf area
  index estimation in temperate deciduous forests from leaf-on and leaf-off
  airborne LiDAR data}}.
\newblock \bibinfo{journal}{Remote Sensing of Environment}
  \bibinfo{volume}{240}.
\newblock \DOIprefix\doi{10.1016/j.rse.2020.111696}.

\end{thebibliography}

\end{document}